\documentclass[10pt,twocolumn,letterpaper]{article}

\usepackage{cvpr}              

\usepackage[accsupp]{axessibility}  
\usepackage{graphicx}
\usepackage{amsmath}
\usepackage{amssymb}
\usepackage{booktabs}
\usepackage{color}
\usepackage{enumitem}
\usepackage{verbatim}
\usepackage{bbm}
\usepackage{multirow}
\usepackage{makecell}
\usepackage{array}
\usepackage{multicol}
\usepackage{multirow}
\usepackage{booktabs}
\usepackage{threeparttable}
\usepackage{xcolor,colortbl}
\usepackage{subcaption}
\usepackage{bm}
\usepackage{bbding}

\def\etal{\textit{et al}.}

\definecolor{Gray}{gray}{0.85}
\definecolor{White}{gray}{1.0}
\newcolumntype{a}{>{\columncolor{Gray}}c}
\newcolumntype{z}{>{\columncolor{White}}c}

\usepackage[pagebackref,breaklinks,colorlinks]{hyperref}

\usepackage[capitalize]{cleveref}

\newif\ifshowcomments
\showcommentsfalse

\ifshowcomments

\newcommand{\TODO}[1]{{\color{red}{[TODO: #1]}}}

\else
\newcommand{\TODO}[1]{}
\newcommand{\revised}[1]{}
\newcommand{\phil}[1]{}
\fi

\newcommand{\ignore}[1]{}
\def\etal{\textit{et al}.}
\def\ie{\textit{i.e.}}
\def\eg{\textit{e.g.}}

\definecolor{Gray}{gray}{0.85}
\definecolor{White}{gray}{1.0}
\newcolumntype{a}{>{\columncolor{Gray}}c}
\newcolumntype{z}{>{\columncolor{White}}c}

\def\cvprPaperID{2742} 
\def\confName{CVPR}
\def\confYear{2023}

\begin{document}

\title{Curricular Contrastive Regularization for Physics-aware Single Image Dehazing}
\author{Yu Zheng$^1$, Jiahui Zhan$^1$, Shengfeng He$^{2}$, Junyu Dong$^1$, Yong Du$^1$\thanks{Corresponding author (csyongdu@ouc.edu.cn).}\\
$^1$ College of Computer Science and Technology, Ocean University of China\\
$^2$ School of Computing and Information Systems, Singapore Management University
}

\maketitle


 \begin{abstract}
Considering the ill-posed nature, contrastive regularization has been developed for single image dehazing, introducing the information from negative images as a lower bound. However, the contrastive samples are non-consensual, as the negatives are usually represented distantly from the clear (\ie, positive) image, leaving the solution space still under-constricted. Moreover, the interpretability of deep dehazing models is underexplored towards the physics of the hazing process. In this paper, we propose a novel curricular contrastive regularization targeted at a consensual contrastive space as opposed to a non-consensual one. Our negatives, which provide better lower-bound constraints, can be assembled from 1) the hazy image, and 2) corresponding restorations by other existing methods. Further, due to the different similarities between the embeddings of the clear image and negatives, the learning difficulty of the multiple components is intrinsically imbalanced. To tackle this issue, we customize a curriculum learning strategy to reweight the importance of different negatives. In addition, to improve the interpretability in the feature space, we build a physics-aware dual-branch unit according to the atmospheric scattering model. With the unit, as well as curricular contrastive regularization, we establish our dehazing network, named C$^2$PNet. Extensive experiments demonstrate that our C$^2$PNet significantly outperforms state-of-the-art methods, with extreme PSNR boosts of 3.94dB and 1.50dB, respectively, on SOTS-indoor and SOTS-outdoor datasets. Code is available at \url{https://github.com/YuZheng9/C2PNet}.
\end{abstract}

 \section{Introduction}
As a common atmospheric phenomenon, haze noticeably degrades the quality of photographed images, severely limiting the performance of subsequent high-level visual tasks such as vehicle re-identification~\cite{chen2022sjdl} and scene understanding~\cite{sakaridis2018model}. Similar to the emergence of other image restoration task solvers~\cite{wen2019single,zhang2019fast,du2020blind,du2023dsdnet}, valid image dehazing techniques are required for handling vision-based applications.
\begin{figure}[t]
	\includegraphics[width=\linewidth]{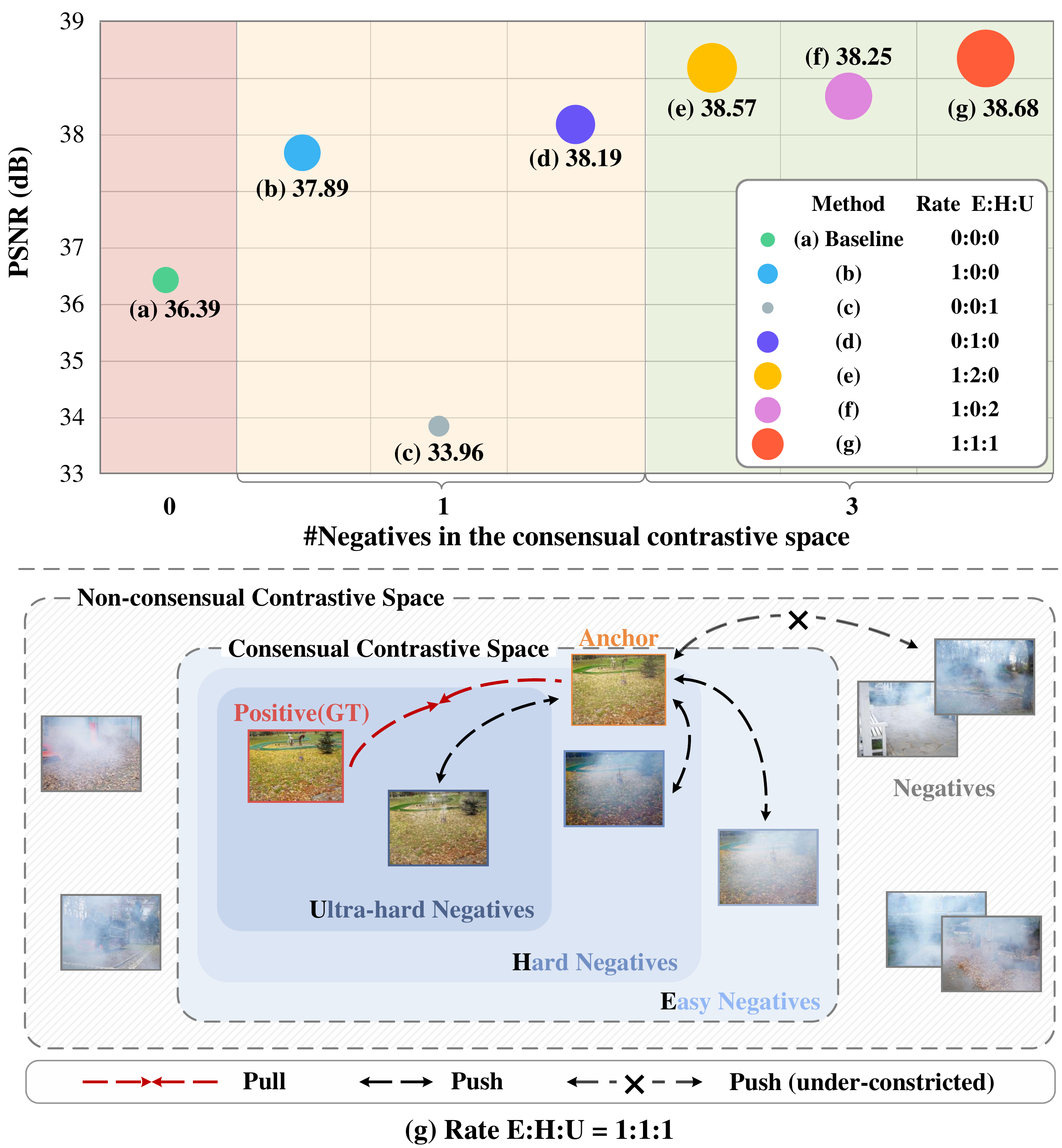}
	\captionsetup{justification=justified}
	\caption{Upper panel: Examination for contrastive regularization based on three difficulty levels of the negatives in the consensual contrastive space. Lower panel: Illustration of contrastive samples in the consensual and non-consensual spaces.}
	\label{fig:teaser}\vspace{-6mm}
\end{figure}

\vspace{-4mm}Deep learning based methods have achieved tremendous success in single image dehazing and can be roughly categorized into two classes: physics-free methods~\cite{chen2019gated, hong2020distilling,dong2020multi,guo2022image} and physics-aware methods~\cite{ren2016single,cai2016dehazenet,dong2020physics,chen2021psd}. Regarding the former, most of them usually use ground-truth images with predicted restorations to enforce L1/L2 distance-based consistency and also involve various regularizations~\cite{zhang2018densely,liu2019griddehazenet} as additional constraints to cope with the ill-posed property. Notice that all of those regularizations ignore the information from negative images as a lower bound, contrastive regularization (CR)~\cite{wu2021contrastive} is proposed to introduce different hazy images as negatives and the ground-truth image as the positive and further uses contrastive learning~\cite{He2020,hadsell2006dimensionality} to guarantee a closed solution space. Moreover, it is shown that better performances can be achieved when using more negatives since diverse degraded patterns are included as cues. However, the issue is that the contents of those negatives are distinct from the positive, and their embeddings may be too distant, leaving the solution space still under-constricted. 

To remedy this issue, a natural idea is to use the negatives in the \textit{consensual} contrastive space\footnote{In this space, the contents of the negatives are identical to the positive sample, except for the haze distribution. Here, we use the terms (non-)consensual contrastive space and (non-)consensual space interchangeably, and a negative in the consensual space is denoted as a consensual negative.} (see the lower panel in Fig.~\ref{fig:teaser}) as better lower-bound constraints, which can be easily assembled from the hazy input and the corresponding restorations by other existing methods. In such cases, the negatives can be ``closer'' to the positive than those in the non-consensual space since the diversity of such negatives is more associated with the haze (or haze residue) rather than any other semantics. However, an intrinsic dilemma arises when the embedding of a negative is too close to that of the positive, as its pushing force to an anchor (\ie, the prediction) may cancel out the pulling force of the positive. Such a learning difficulty can confuse the anchor to move towards the positive, especially in the early training stage. 

This intuition is further examined in the upper panel of Fig.~\ref{fig:teaser}. We use FFA-Net~~\cite{qin2020ffa} as baseline (row (a)) and SOTS-indoor~\cite{li2018benchmarking} as the testing dataset to explore the impact of the negatives in the consensual space with diverse difficulty. Specifically, we define the difficulty of the negatives into three levels: easy (E), hard (H), and ultra-hard (U). We adopt the hazy input as the easy negative, and use a coarse strategy to distinguish between the latter two types, \ie, whether the PSNR of the negative is greater than 30. First, in the single-negative case (row (b)-(d)), an interesting finding is that using a hard sample as negative achieves the best performance compared to the other two settings, and using an ultra-hard negative is even worse than the baseline. This reveals that a ``close'' negative has the potential to promote the effectiveness of the dehazing model, but not the closer the better due to the learning difficulty. While in the multi-negative case\footnote{We give each negative the same weight in the regularization under this case, and we omit the cases of E=0, which would drastically decrease the performance. We will discuss the reason for this in Sec.~\ref{section:A}.} (row (e)-(g)), we have observed that comprehensively covering negatives with different difficulty levels, including ultra-hard samples, can lead to the best performance. It implies the negatives at different difficulty levels can all contribute to the training phase. These observations motivate us to explore how to wisely arrange the multiple negative pairs in a consensual space into the CR during training.

Moving on to the realm of physics-aware deep models, most of them utilize the atmospheric scattering model~\cite{mccartney1976optics,nayar1999vision} in the raw space, without fully exploring the beneficial feature-level information. PFDN~\cite{dong2020physics} is the only work that attempts to express the physics model as a basic unit in the network. The unit is designed as a shared structure to predict the latent features corresponding to the atmospheric light and transmission map. Nevertheless, the former is usually assumed to be homogeneous while the latter is non-homogeneous, and thus their features cannot be approximated in the same way. Therefore, it is still an open problem how to accurately realize the interpretability of the feature space of the deep network using the physics model, which is another aspect we are interested in.

In this paper, we propose a curricular contrastive regularization using hazy or restored images as negatives in the consensual space for image dehazing to address the first issue. Informed by our analysis, which suggests that the difficulty of consensual negatives can impact the effectiveness of the regularization, we present a curriculum learning strategy to arrange these negatives to mitigate learning ambiguity. Specifically, we split the negatives into three types (\ie, easy, hard, and ultra-hard) and assign different weights to corresponding negative pairs in CR. Meanwhile, the difficulty levels of the negatives are dynamically adjusted as the anchor moves towards the positive in the representation space during training. In this way, the proposed regularization can facilitate the dehazing models to be stably optimized in a more compact solution space.

We propose a physics-aware dual-branch unit (PDU) regarding the second issue. The PDU approximates the features corresponding to the atmospheric light and the transmission map in dual branches, respectively considering the physical characteristics of each factor. The features of the latent clear image can thus be synthesized more precisely in line with the physics model. Finally, we establish C$^2$PNet, our dehazing network that deploys PDUs into a cascaded backbone with curricular contrastive regularization.

In summary, our key contributions are as follows:
\begin{itemize}
	\item \vspace{-2mm}We propose a novel C$^2$PNet for haze removal that employs curricular contrastive regularization and enforces physics-based prior in the feature space. Our method outperforms SOTAs in both synthetic and real-world scenarios. In particular, we achieve significant PSNR boosts of 3.94dB and 1.50dB on the SOTS-indoor and SOTS-outdoor datasets, respectively.
	
	\item \vspace{-2mm}The proposed regularization adopts a unique consensual negative-based approach for dehazing and incorporates a self-contained curriculum learning strategy that dynamically calibrates the priority and difficulty levels of the negatives. It is also proven to enhance the performance of SOTAs as a generalized regularization technique, surpassing previous related strategies. 
	
	\item \vspace{-2mm}With careful consideration of the characteristics of factors involved, we built the PDU based on an unprecedented expression of the physics model. This innovative design promotes feature transmission and extraction in the feature space, guided by physics priors.
	
\end{itemize}

 \section{Related Work}
\label{related}
\begin{figure*}[t]
	\center
	\includegraphics[width=\linewidth]{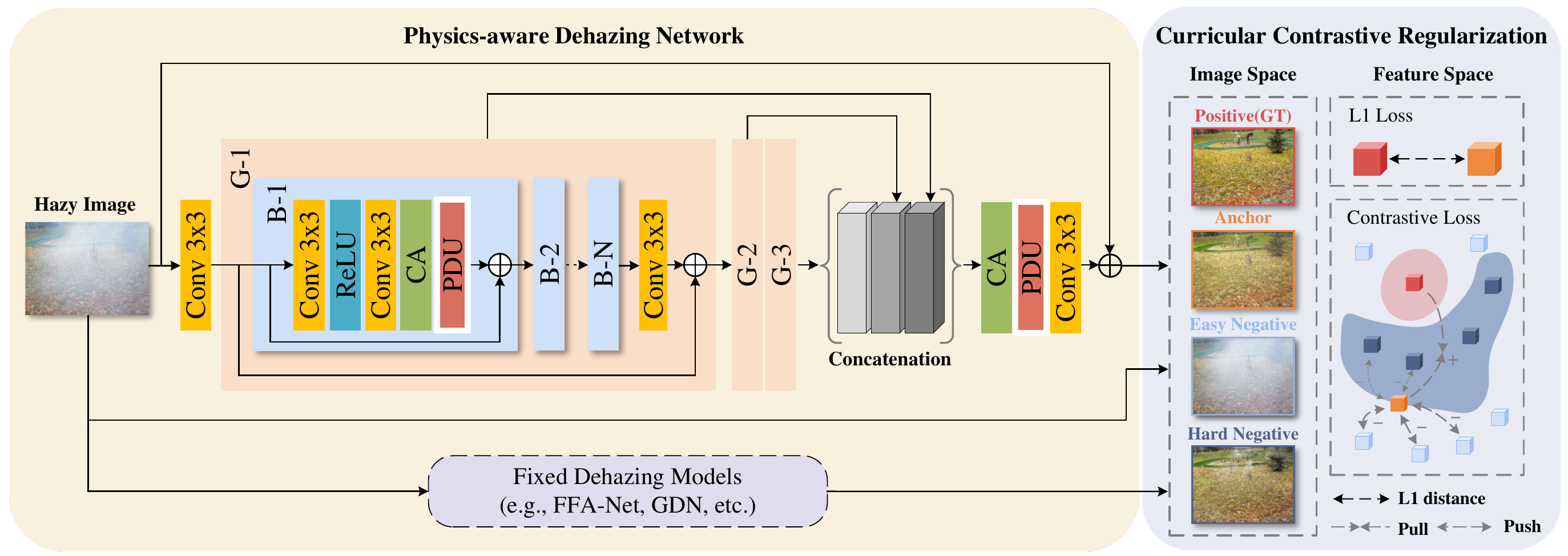}
	\caption{Illustration of our C$^2$PNet for single image dehazing.}
	\label{fig:method}\vspace{-4mm}
\end{figure*}
\textbf{Single Image Dehazing.} 
Traditional single image dehazing methods are mainly based on an atmospheric scattering model~\cite{mccartney1976optics}. They focus on designing hand-crafted priors such as the dark channel prior~\cite{he2010single} and color attenuation prior~\cite{zhu2015fast}. However, these priors may not be powerful enough to characterize complex scenes in practice. Early learning-based methods~\cite{cai2016dehazenet,ren2016single} use deep neural networks to predict the transmission map and atmospheric light in the physics model to obtain a latent clear image. However, inaccuracies in the estimations may accumulate, hindering the reliable inference of the haze-free image. With the advent of large haze datasets~\cite{li2018benchmarking}, data-driven methods~\cite{chen2021psd,liu2021synthetic,qin2020ffa,guo2022image} have been developed rapidly. FFANet~\cite{qin2020ffa} introduces feature attention (FA) blocks that leverage both channel and pixel attention to improve haze removal. DeHamer~\cite{guo2022image} combines CNN and Transformer for image dehazing, which can aggregate long-term attention in Transformer and local attention in CNN features. Note that these methods do not consider the physics of the hazing process. Further, Dong \etal~propose a feature dehazing unit (FDU)~\cite{dong2020physics} derived based on the physics model. To the best of our knowledge, this work is the only one that considers the physics model in the feature space, avoiding the cumulative errors that occur in the raw space. However, FDU uses a shared structure to predict those unknown factors without considering their different physical characteristics. To solve this problem, we re-understand the physics model and construct a novel physics-aware dual-branch unit for image dehazing.

\textbf{Contrastive Learning.} 
In recent, contrastive learning has been broadly employed in high-level visual tasks~\cite{grill2020bootstrap,He2020,chen2020simple,guo2022hcsc}. The idea behind contrastive learning is to pull an anchor point closer to a positive point while simultaneously pushing it away from a negative point through a contrastive loss. However, there are only a few works that have applied contrastive learning to low-level vision problems. CR~\cite{wu2021contrastive} is one of the representative works, which introduces the concept of negative points for image dehazing. By considering the negative information as a lower bound of the solution space, CR can exploit both positive and negative information for training. However, most of the negatives are non-consensual and thus distantly represented from the positive, resulting in an under-constrained solution space. We aim to solve this issue with a novel curricular contrastive regularization approach that uses consensual negatives.

\textbf{Curriculum Learning.}
Inspired by the cognitive systems of humans, Elman~\cite{elman1993learning} emphasizes the importance of starting small in neural network training, which may be considered a prototype of curriculum learning. Later, Bengio \etal ~\cite{bengio2009curriculum} formally propose the curriculum learning strategy to arrange the training samples according to their difficulty. Nowadays, curriculum learning has been successfully applied to various cases including vision and language tasks~\cite{schroff2015facenet,korbar2018cooperative,duan2020curriculum,zhang2021flexmatch}. Building on our analysis that different consensual negatives exhibit varying learning difficulty, the question arises of how to arrange these samples during training. We propose to solve this issue via a self-contained curriculum learning strategy.

\section{Method}
\label{section:A}
\subsection{Overview}
Our goals are two-fold: 1) to promote the interpretability of the feature space for haze removal and 2) to establish a more concise solution space using of contrastive samples. Fig.~\ref{fig:method} illustrates the detailed structure of our C$^2$PNet. To achieve our first goal, we design a physics-aware dual-branch unit that is derived from the atmospheric scattering model. Regarding our second aim, we tailor a contrastive regularization using consensual negatives, along with a self-contained curriculum learning strategy to deal with the learning difficulty. Note that our curricular contrastive regularization is network-agnostic, making it applicable to other dehazing networks.
\subsection{Physics-aware Dual-branch Unit}
\begin{figure}[t]
	\center
	\includegraphics[width=\linewidth]{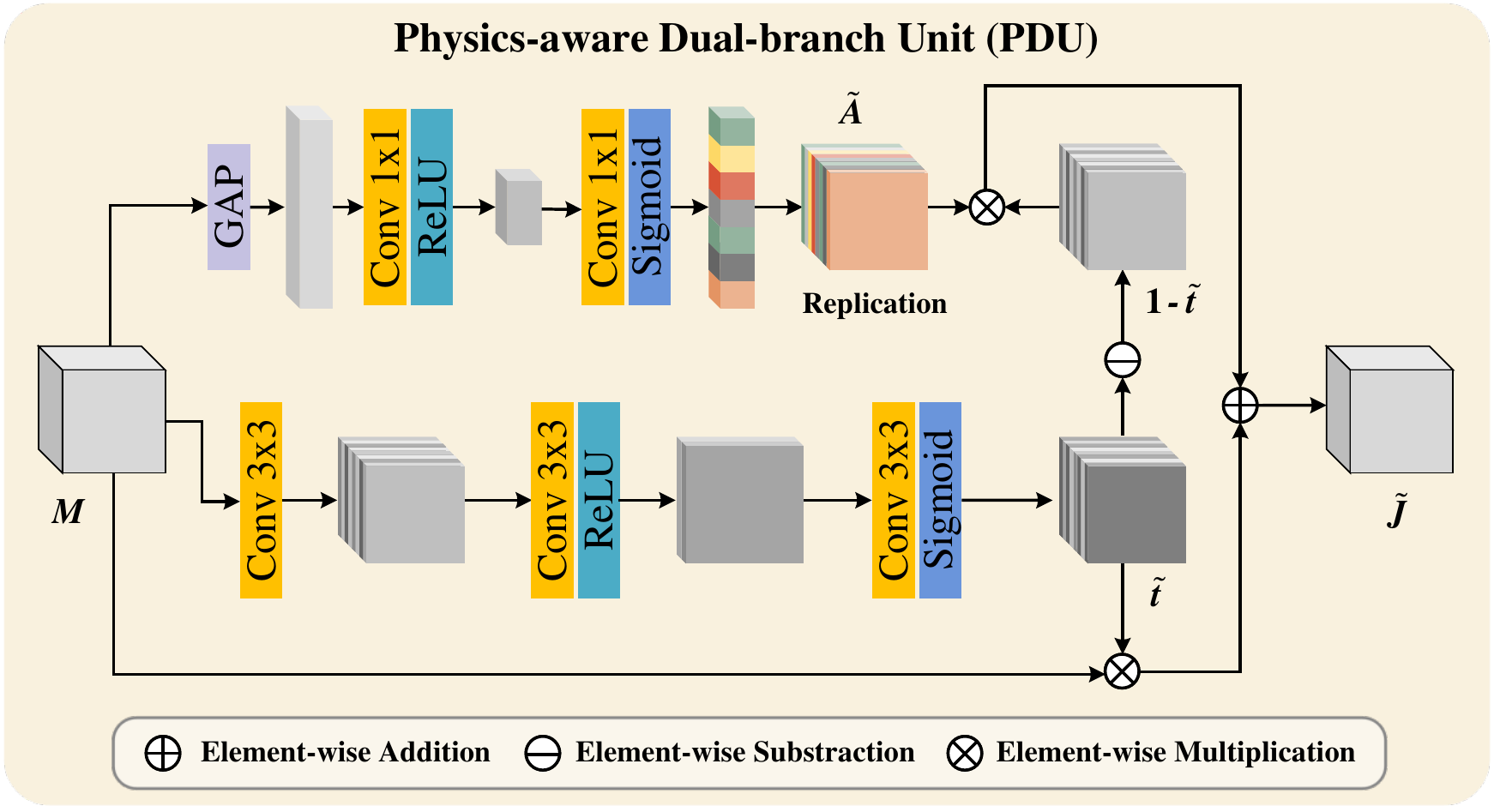}
	\caption{The architecture of the proposed PDU.}
	\label{fig:block}\vspace{-6mm}	
\end{figure}

The atmospheric scattering model is commonly used to describe the formation of a hazy image $I$. It can be mathematically formulated as $I(x)=T(x)J(x)+(1-T(x))A$, where $J$ represents the clear image, $T$ is the transmission map, $A$ indicates the atmospheric light, and $x$ denotes the index of pixels. As both $T$ and $A$ are unknown, haze removal is a highly ill-posed problem. Raw space based methods directly estimate the two unknown factors, which can easily lead to cumulative errors. In contrast, imposing physics priors in the feature space can encourage the interpretability that aligns with the hazing process, without relying on the ground truths of $T$ and $A$. Inspired by FDU~\cite{dong2020physics}, we propose a physics-aware dual-branch Unit (PDU) that is derived from the physics model in the feature space, as shown in Fig.~\ref{fig:block}. 

To begin with, we reformulate the physics model to represent the clear image $J$ as follows:
\begin{equation}
	\begin{aligned}
		J(x)&=I(x)\frac{1}{T(x)}+A(1-\frac{1}{T(x)})\\
		&=I(x)\frac{1}{T(x)}+A-A\frac{1}{T(x)}.
	\end{aligned}
    \label{equ:scatter}
\end{equation}
Then extracting features via kernel $k$, Eq.~\eqref{equ:scatter} can be reformulated as follows:
\begin{equation}
	k\circledast J=k\circledast(I\odot\frac{1}{T})+k\circledast A-k\circledast(A\odot\frac{1}{T}), 
	\label{equ:conv}
\end{equation}
where $\circledast$ indicates the convolution operator and $\odot$ denotes the Hadamard product.  Consequently, we respectively introduce the matrix-vector forms of $k$, $J$, $I$, $A$, $\frac{1}{T}$, \ie, $\bm{K}$, $\bm{J}$, $\bm{I}$, $\bm{A}$ and $\bm{D}$, and Eq.~\eqref{equ:conv} can be rewritten as 
\begin{equation}
	\bm{KJ}=\bm{KDI}+\bm{KA}-\bm{KDA}.
\end{equation}
Such a reformulation can be given by a few steps of algebra operations. Note that the diagonal vector of the diagonal matrix $\bm{D}$ corresponds to the vectorized form of $\frac{1}{T}$.

Next, we can decompose the matrix $\bm{KD}$ into a product of two matrices $\bm{QK}$. As $\bm{K}$, $\bm{D}$ and $\bm{Q}$ are all unknown, implementing this decomposition can be indicated as solving an underdetermined system of equations, which can guarantee the existence of $\bm{Q}$. And then, we have
\begin{equation}	
	\bm{KJ}=\bm{Q}(\bm{KI})+\bm{KA}-\bm{Q}(\bm{KA}).
	\label{equ:Q}		
\end{equation}

We can denote $\tilde{\bm{A}}$ as an approximation of the features $\bm{KA}$ that correspond to the atmospheric light and $\tilde{\bm{t}}$ as an approximation of $\bm{Q}$, which is associated with the transmission map. Furthermore, $\bm{KI}$ and $\bm{KJ}$ can be viewed as the extracted features of a hazy image and its corresponding clear image, respectively. Based on Eq.~\eqref{equ:Q}, and assuming that the channel number of the features $\tilde{\bm{t}}$ matches that of the input features $\bm{M}$, we can calculate the physics-aware features $\tilde{\bm{J}}$ by 
\begin{equation}	
	\begin{aligned}
		\tilde{\bm{J}}&=\bm{M}\odot\tilde{\bm{t}}+\tilde{\bm{A}}-\tilde{\bm{A}}\odot\tilde{\bm{t}}\\		&=\bm{M}\odot\tilde{\bm{t}}+\tilde{\bm{A}}(\bm{1}-\tilde{\bm{t}}),
	\end{aligned}
    \label{equ:final}
\end{equation}
where $\bm{1}$ indicates a matrix whose elements are all ones. 

Note that the second term on the right-hand side of Eq.~\eqref{equ:final} involves a synergistic action between $\tilde{\bm{A}}$ and $\tilde{\bm{t}}$ that is ignored by FDU. Then we can explicitly build the PDU based on Eq.~\eqref{equ:final}. One branch in PDU (see the upper part of Fig.~\ref{fig:block}) is used to produce $\tilde{\bm{A}}$. As the atmospheric light is usually assumed to be homogeneous, we use global average pooling (GAP($\cdot$)) to eliminate unnecessary information in the feature space. And $\tilde{\bm{A}}$ is produced by
\begin{equation}	
	\tilde{\bm{A}}=H(\sigma(\textrm{Conv}^N(\textrm{ReLU}(\textrm{Conv}^{\frac{N}{8}}(\textrm{GAP}(\bm{M})))))),		
\end{equation}
where $\sigma(\cdot)$ is the Sigmoid function, $H(\cdot)$ denotes a replication operation, $\textrm{Conv}^N(\cdot)$ is the convolutional layer with $N$ kernels, and $N$ is set to 64.

On the other hand, we cannot apply GAP$(\cdot)$ for the approximation of $\bm{Q}$ due to a loss of information, as the transmission map is non-homogeneous. Therefore, in the lower branch in Fig.~\ref{fig:block}, we choose to extract $\tilde{\bm{t}}$ using a sequence of convolutional layers, which is given by
\begin{equation}	
	\tilde{\bm{t}}=\sigma(\textrm{Conv}^N(\textrm{ReLU}(\textrm{Conv}^{\frac{N}{8}}(\textrm{Conv}^{N}(\bm{M}))))).		
\end{equation}

With the proposed PDU, interpretable features $\tilde{\bm{J}}$ can be generated from the input features $\bm{M}$ for restoring hazy images. Unlike FDU, which uses a shared structure with GAP$(\cdot)$ to predict latent features that are simultaneously correlated to both $T$ and $A$, the PDU attentively incorporates the corresponding physical characteristics of these two factors. This approach allows for more useful features to be estimated in a dual interactive paradigm. 

\subsection{Curricular Contrastive Regularization} 
\begin{figure}[t]
	\center
	\includegraphics[width=\linewidth]{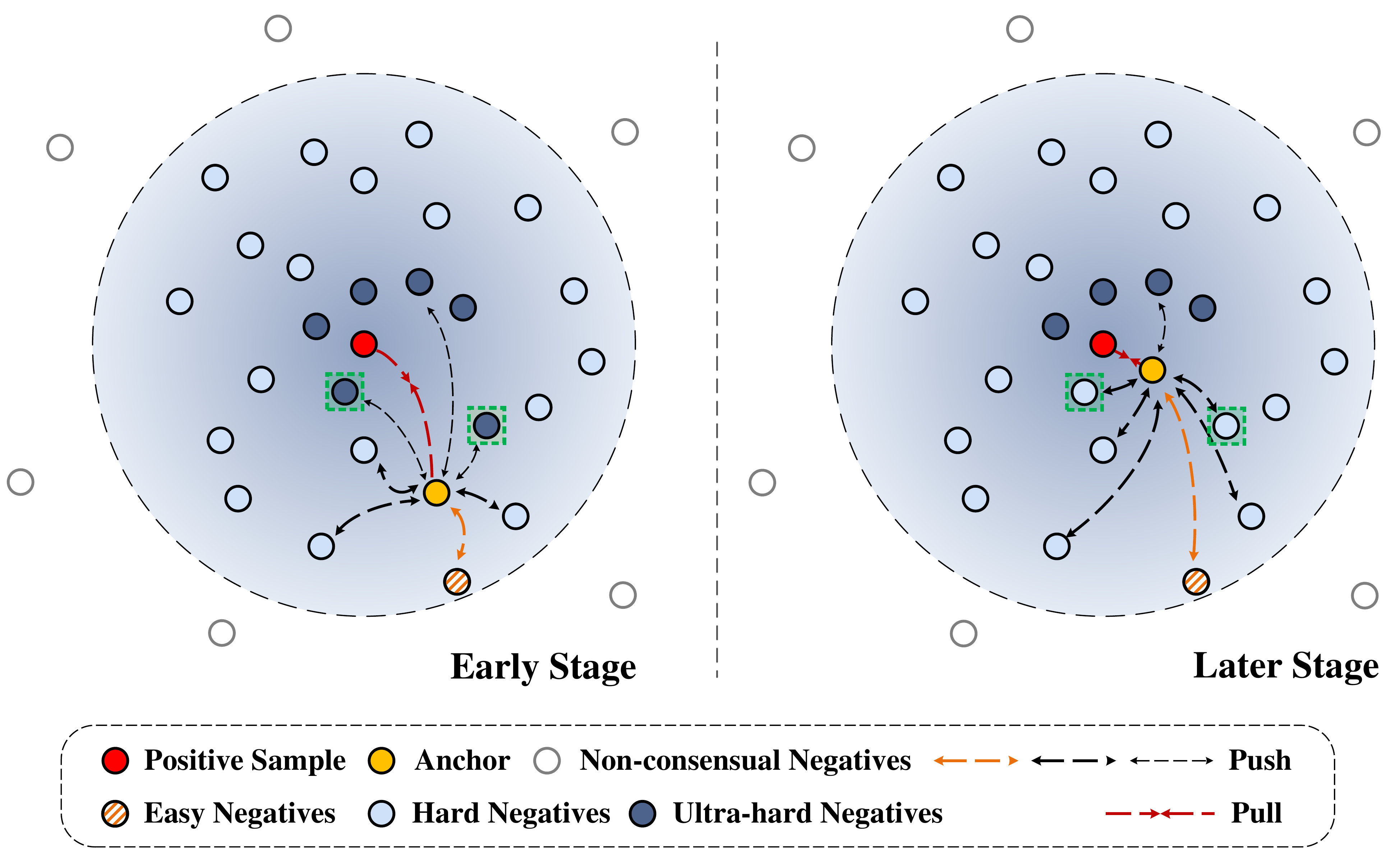}
	\caption{Illustration of curricular contrastive regularization.}
	\label{fig:CC}\vspace{-4mm}
\end{figure}

Regarding the canonical contrastive regularization for image dehazing, the anchor is the recovered result by the dehazing network, the positive is the ground truth, and the negatives include a hazy input and multiple hazy images that are non-consensual with the positive. The target of this regularization $R$ is to minimize the L1 distance between the embeddings of the anchor and the positive while maximizing their distance from the negatives, which is given by
\begin{equation}
	R=\sum_{i=1}^n\xi_i\frac{||V_i(J)-V_i(f(I,\theta))||_1}{\sum_{q=1}^r||V_i(U_q)-V_i(f(I,\theta))||_1+E_i},		
\end{equation}
where $E_i=||V_i(I)-V_i(f(I,\theta))||_1$, $f(\cdot,\theta)$ indicates the dehazing network with parameters $\theta$, $V_i(\cdot), i=1,2,\cdots,n$ extracts the $i$th hidden features from the pre-trained VGG-19~\cite{simonyan2014very}, the number of non-consensual negatives $\{U_q\}$ is $r$, and $\{\xi_{i}\}$ is the set of hyperparameters. As illustrated in Fig.~\ref{fig:CC}, the introduced contrast between the anchor and non-consensual negatives cannot provide a satisfactory lower bound of the solution space. The non-consensual negatives are typically distantly located from the positive, leading to an under-constricted solution space that limits the quality of the restorations.

Based on our analysis of Fig.~\ref{fig:teaser}, we propose a novel contrastive regularization for haze removal that utilizes negatives in the consensual space, which can be restored results from other dehazing models. Our straightforward aim is to push the anchor far away from better-quality negatives. However, two critical problems arise: 1) how to define the difficulty of different negatives and 2) how to arrange these negatives according to their difficulty during training. 

To solve both issues, we incorporate a curriculum learning strategy into contrastive regularization. We define the difficulty of the negatives into three levels: easy, hard, and ultra-hard. For easy negative, we use the hazy input consistently. The difficulty levels of the other negatives are dynamically determined during training. Specifically, we measure the average PSNR performance of the network before every epoch begins. In the $t$th epoch, a negative is defined as an ultra-hard sample when its PSNR is higher than the network performance, or as a hard negative otherwise. 

To properly arrange these negatives, we weigh them differently according to their difficulty levels. First, the weight of easy negative is fixed and largest. This is because although hard and ultra-hard negatives may contribute to a more compact solution space, they can also cause learning ambiguity. To ensure that the resultant force is towards the positive such that the anchor is shifted in the desired direction, we give the easy negative a weight that is large enough. In practice, we set this weight to the number of the non-easy negatives $z$. Second, the weight of a non-easy negative $S_q$ at the $t$th epoch is defined as follows: 
\begin{small}
\begin{equation}
	W_t(S_q) = \left\{
	\begin{array}{rcl}
		1+\gamma, &\textrm{avgPSNR}(f(\{I_g\},\theta_{t-1}))\geq \textrm{PSNR}(S_q),\\
		1-\gamma, & \textrm{otherwise},\\
	\end{array} \right. 
	\label{equ:beta}
\end{equation}
\end{small}
where $\{I_g\}$ denotes the hazy input dataset, $q=1,2,\cdots,z$ is the index of the non-easy negatives, and $\gamma$ is a hyperparameter. The weights of the hard and the ultra-hard negatives are set to $1+\gamma$ and $1-\gamma$, respectively. This means that the weight of a hard negative is larger than that of an ultra-hard negative, allowing the hard negative to provide a greater force and alleviating the potential learning ambiguity. Furthermore, the flexibility of this strategy in determining the difficulty levels enables ultra-hard negatives to become hard ones in the later stage of training (see Fig.~\ref{fig:CC}). This makes sense because as the quality of the anchor improves, the ambiguity caused by ultra-hard samples is reduced, and their importance should be strengthened. In this way, the hard and ultra-hard negatives can be viewed as better lower bounds for effectively constraining the solution space. Then, our curricular contrastive regularization $R^*$is formulated as follows:
\begin{small}
\begin{equation}
	R^*=\sum_{i=1}^n\xi_i\frac{||V_i(J)-V_i(f(I,\theta))||_1}{\sum_{q=1}^zW_t(S_q)||V_i(S_q)-V_i(f(I,\theta))||_1+z\cdot E_i}.
	\label{equ:R}	
\end{equation}
\end{small}

Finally, our total objective $\cal L$, which consists of an L1 norm based fidelity term and our contrastive curricular regularization, is given by
\begin{equation}
	{\cal L}=||J-f(I,\theta)||_1+\lambda R^*.	
\end{equation}

\subsection{Network Architecture}
\begin{table*}[t]
	\caption{Quantitative Evaluations with the state-of-the-art methods on the synthetic and real-world datasets.}
	\centering
	\small
	\begin{tabular}{c||c||c|c||c|c||c|c||c|c||c}
		\toprule
		\multirow{2}*{Method} &\multirow{2}*{Venue\&Year}&\multicolumn{2}{c||}{SOTS-indoor} &\multicolumn{2}{c||}{SOTS-outdoor} &\multicolumn{2}{c||}{Dense-Haze} &\multicolumn{2}{c||}{NH-Haze2} &\multirow{2}*{\#Params} \\
		\cmidrule(lr){3-4}
		\cmidrule(lr){5-6}
		\cmidrule(lr){7-8}
		\cmidrule(lr){9-10}		
		&&PSNR&SSIM&PSNR&SSIM&PSNR&SSIM&PSNR&SSIM&\\
		\midrule
		DCP~\cite{he2010single}&TPAMI2010&16.62&0.8179&19.13&0.8148&11.01&0.4165&11.68&0.6475&-\\
		
		DehazeNet~\cite{cai2016dehazenet}&TIP2016&21.14&0.8472&22.46&0.8514&9.48&0.4383&11.77&0.6217&0.01M\\
		
		AODNet~\cite{li2017aod}&ICCV2017&19.06&0.8504&20.29&0.8765&12.82&0.4683&12.33&0.6311&0.002M\\	
		
		DM2F-Net~\cite{Deng2019}&ICCV2019&34.29&0.9728&34.50&0.9815&14.99&0.5640&20.46&0.8217&92.14M\\
		
		GCANet~\cite{chen2019gated}&WACV2019&30.06&0.9596&22.76&0.8887&12.62&0.4208&18.79&0.7729&0.70M\\
		
		GDN~\cite{liu2019griddehazenet}&ICCV2019&32.16&0.9836&30.86&0.9819&14.96&0.5326&19.26&0.8046&0.96M\\	
		
		MSBDN~\cite{dong2020multi}&CVPR2020&32.77&0.9812&34.81&0.9857&15.13&0.5551&20.11&0.8004&31.35M\\	
		
		FFA-Net~\cite{qin2020ffa}&AAAI2020&36.39&0.9886&33.57&0.9840&12.22&0.4440&20.00&0.8225&4.46M\\	
		
		AECR-Net~\cite{wu2021contrastive}&CVPR2021&37.17&0.9901&-&-&15.80&0.4660&20.68&0.8282&2.61M\\
		
		MAXIM-2S~\cite{tu2022maxim}&CVPR2022&38.11&0.9908&34.19&0.9846&-&-&-&-&14.1M\\	
		
		DeHamer~\cite{guo2022image}&CVPR2022&36.63&0.9881&35.18&0.9860&16.62&0.5602&19.18&0.7939&132.45M\\	
		
		UDN~\cite{hong2022uncertainty}&AAAI2022&38.62&0.9909&34.92&0.9871&-&-&-&-&4.25M\\		
		\midrule
		\textbf{C$^2$PNet}   &&\textbf{42.56}&\textbf{0.9954}&\textbf{36.68}&\textbf{0.9900}&\textbf{16.88}&\textbf{0.5728}&\textbf{21.19}&\textbf{0.8334}&7.17M\\		
		\bottomrule
	\end{tabular}
	\label{tab:quantitative}
\end{table*}
\begin{figure*}[t]
	\centering
	\setlength{\abovecaptionskip}{0cm}
	\setlength{\tabcolsep}{0.05em}
	\setlength{\fboxrule}{1pt}
	\setlength{\fboxsep}{0pt}
	\begin{tabular}{cccccccc}			   		
		PSNR / SSIM& $18.09 /  0.7459 $ & $31.55 / 0.9793$ & $34.41 / 0.9811$ & $36.69 / 0.9838$ & $37.10 / 0.9825$ & $41.20 / 0.9914$ & $\infty / 1$ \\			
		\includegraphics[width=.12\linewidth]{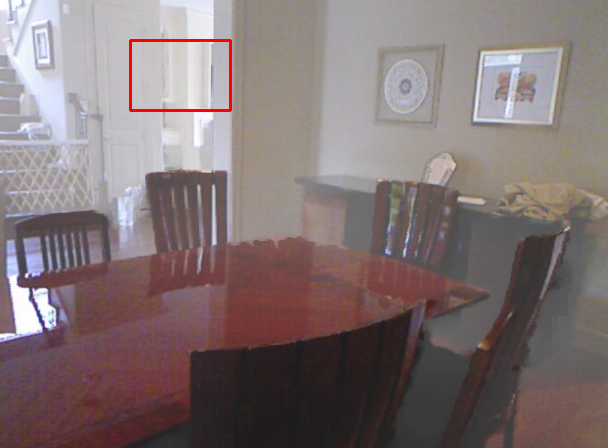} &
		\includegraphics[width=.12\linewidth]{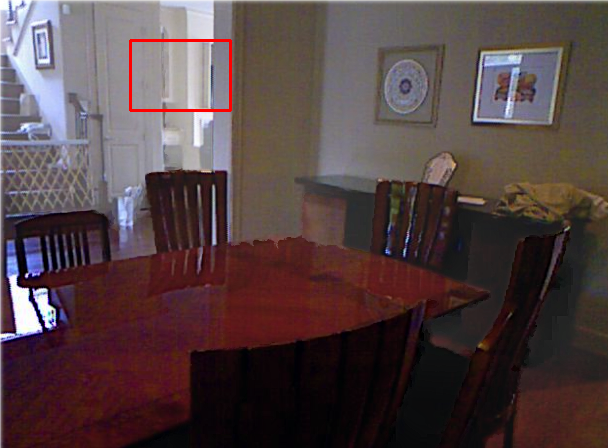} &
		\includegraphics[width=.12\linewidth]{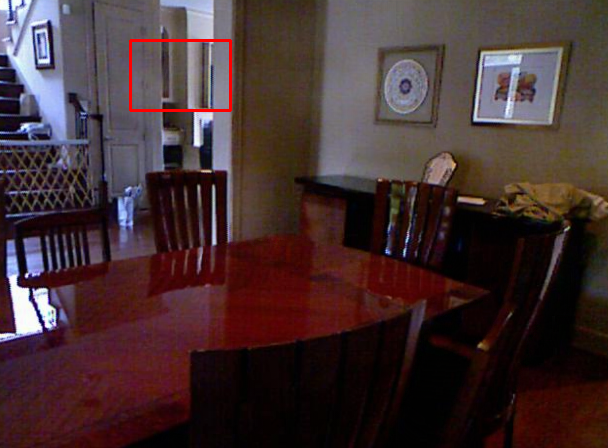} &
		\includegraphics[width=.12\linewidth]{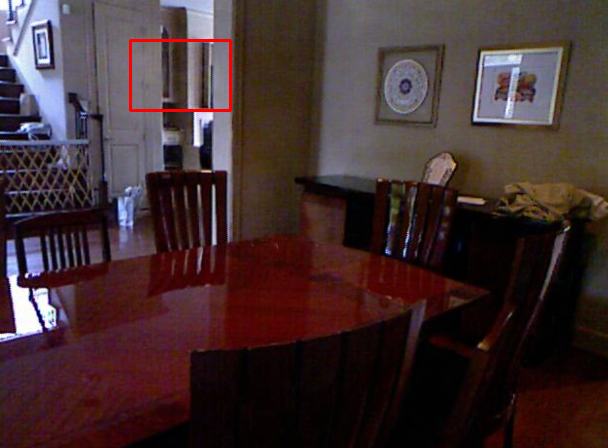} &
		\includegraphics[width=.12\linewidth]{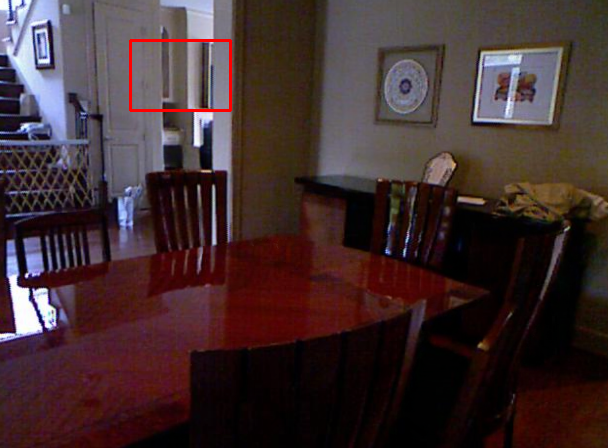} &
		\includegraphics[width=.12\linewidth]{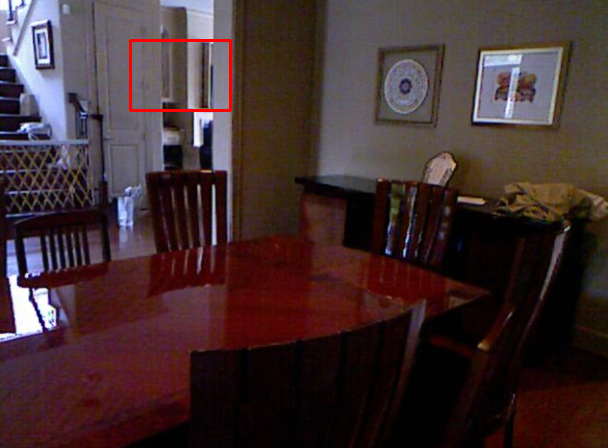} &
		\includegraphics[width=.12\linewidth]{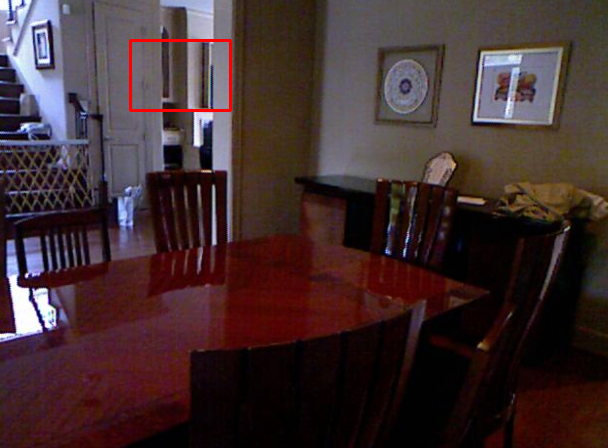}&
		\includegraphics[width=.12\linewidth]{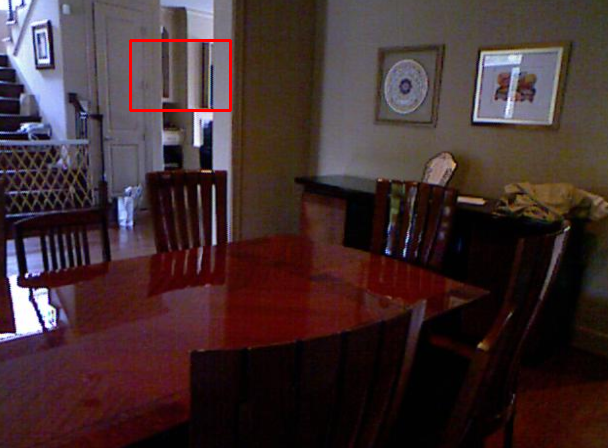}\\	
		\fcolorbox{red}{red}{\includegraphics[width=.117\linewidth]{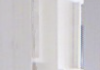}} &
		\fcolorbox{red}{red}{\includegraphics[width=.117\linewidth]{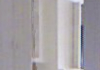}} &
		\fcolorbox{red}{red}{\includegraphics[width=.117\linewidth]{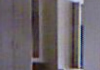}} &
		\fcolorbox{red}{red}{\includegraphics[width=.117\linewidth]{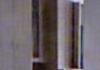}}&
		\fcolorbox{red}{red}{\includegraphics[width=.117\linewidth]{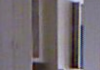}} &
		\fcolorbox{red}{red}{\includegraphics[width=.117\linewidth]{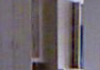}}&
		\fcolorbox{red}{red}{\includegraphics[width=.117\linewidth]{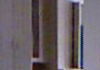}}&
		\fcolorbox{red}{red}{\includegraphics[width=.117\linewidth]{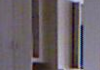}}\\
		Hazy Image &AODNet~\cite{li2017aod}&GDN~\cite{liu2019griddehazenet}&FFA-Net~\cite{qin2020ffa}&MAXIM~\cite{tu2022maxim}&DeHamer~\cite{guo2022image}&C$^2$PNet (Ours)&GT
	\end{tabular}
	\caption{Visual results of SOTS-indoor dataset by different methods. (Zoom in for better view.)
	}
	\label{fig:indoor}
\end{figure*}

Our C$^2$PNet adopts an FFA-Net-like backbone because: 1) FFA-Net has a simple structure that cascades several FA blocks without any other redundant modules, and 2) the FA block is simple and has been proven to be practical. Since the proposed PDU mainly focuses on refining spatial information, we deploy it into each FA block by replacing the PA module. In this way, the features are enforced to conform to the hazing process before being fed into the subsequent module. Note that all other network parameters of C$^2$PNet are identical to those of FFA-Net, except for the PDUs.

 \section{Experiments}
\label{exp}
\begin{figure*}[t]
	\centering
	\setlength{\abovecaptionskip}{0cm}
	\centering
	\setlength{\tabcolsep}{0.05em}
	\setlength{\fboxrule}{1pt}
	\setlength{\fboxsep}{0pt}
	\begin{tabular}{cccccccc}
		PSNR / SSIM& 17.16 / 0.8792  & 23.12 / 0.9598 & 26.64 / 0.9757 & 27.39 / 0.9718 & 29.94 / 0.9758 & 31.10 / 0.9859 & $\infty$ / 1 \\	   		
		\includegraphics[width=.12\linewidth]{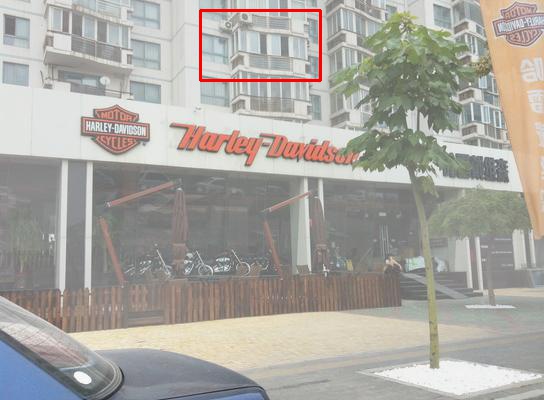} &
		\includegraphics[width=.12\linewidth]{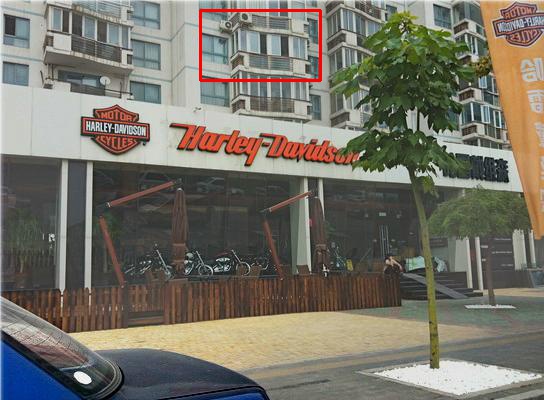} &
		\includegraphics[width=.12\linewidth]{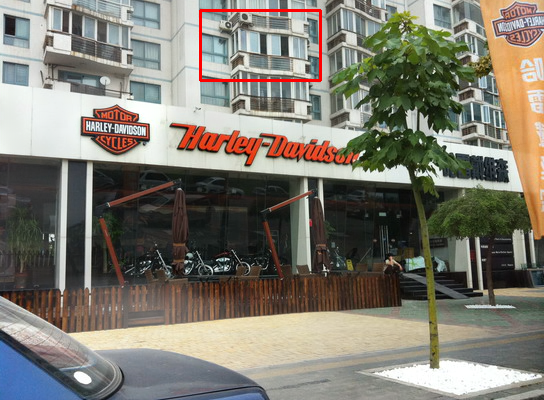} &
		\includegraphics[width=.12\linewidth]{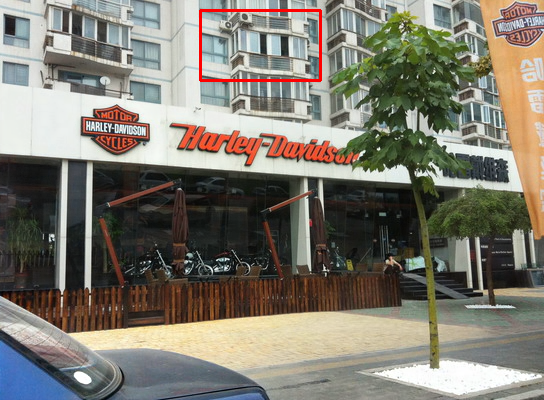} &
		\includegraphics[width=.12\linewidth]{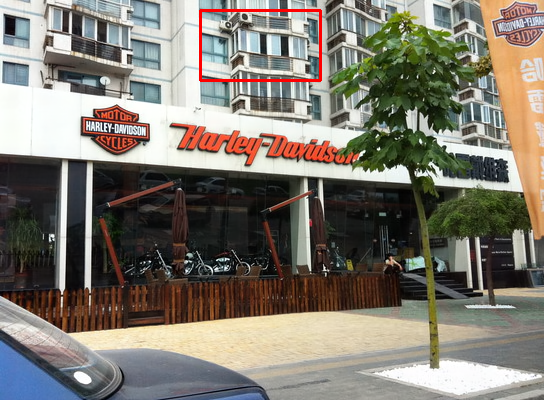} &
		\includegraphics[width=.12\linewidth]{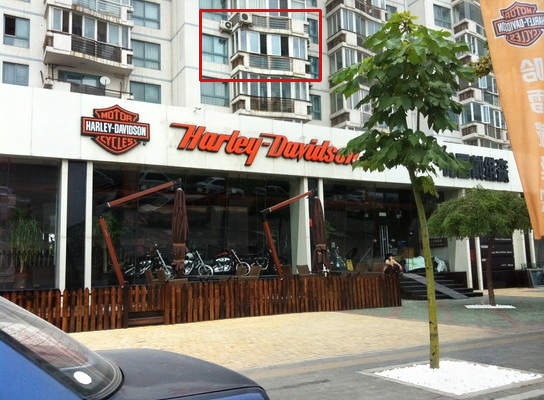} &
		\includegraphics[width=.12\linewidth]{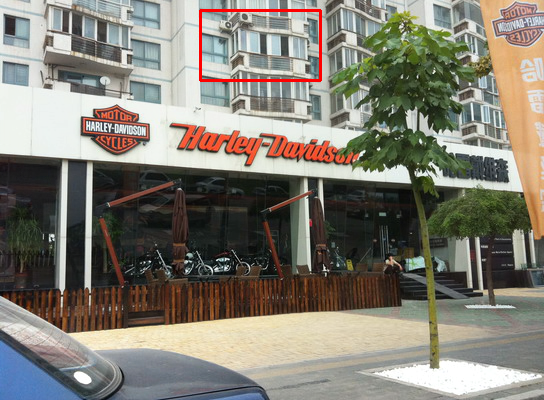}&
		\includegraphics[width=.12\linewidth]{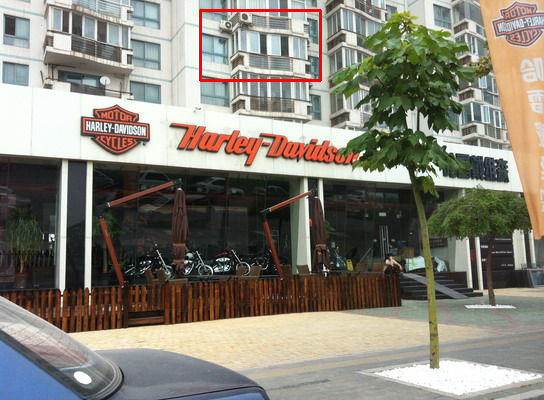}\\	
		\fcolorbox{red}{red}{\includegraphics[width=.117\linewidth]{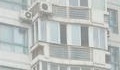}} &
		\fcolorbox{red}{red}{\includegraphics[width=.117\linewidth]{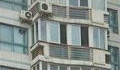}} &
		\fcolorbox{red}{red}{\includegraphics[width=.117\linewidth]{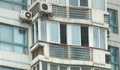}} &
		\fcolorbox{red}{red}{\includegraphics[width=.117\linewidth]{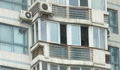}}&
		\fcolorbox{red}{red}{\includegraphics[width=.117\linewidth]{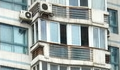}} &
		\fcolorbox{red}{red}{\includegraphics[width=.117\linewidth]{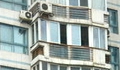}}&
		\fcolorbox{red}{red}{\includegraphics[width=.117\linewidth]{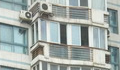}}&
		\fcolorbox{red}{red}{\includegraphics[width=.117\linewidth]{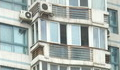}}\\
		Hazy Image &AODNet~\cite{li2017aod}&GDN~\cite{liu2019griddehazenet}&FFA-Net~\cite{qin2020ffa}&MAXIM~\cite{tu2022maxim}&DeHamer~\cite{guo2022image}&C$^2$PNet (Ours)&GT
	\end{tabular}
	\caption{Visual results of SOTS-outdoor dataset by different methods. (Zoom in for better view.)
	}
	\label{fig:outdoor}
\end{figure*}

\subsection{Experimental Settings}
\textbf{Implementation Details.} We implement C$^2$PNet using Pytorch 1.11.0 on an NVIDIA RTX 3090 GPU. Adam optimizer is used with exponential decay rates $\beta_1=0.9$ and $\beta_2=0.999$. The initial learning rate is set to 0.0001 and is scheduled by cosine annealing strategy~\cite{he2019bag}. The batch size is set to 2. We empirically set the penalty parameters $\lambda$ to 0.2, and $\gamma$ to 0.25 for 200 epochs. We follow CR~\cite{wu2021contrastive} that set the L1 distance in Eq.\eqref{equ:R} after the latent features of the 1st, 3rd, 5th, 9th and 13th layers from the fixed pre-trained VGG-19, and their corresponding weights $\xi_i, i=1,\cdots,5$ to $\frac{1}{32},\frac{1}{16},\frac{1}{8},\frac{1}{4},$ and 1, respectively.

\textbf{Datasets.} For fair comparisons, we evaluate the proposed method on synthetic datasets and real-world datasets. RESIDE~\cite{li2018benchmarking} is a widely used benchmark dataset. Among the five subsets, we select ITS and OTS as our training datasets and SOTS-indoor and SOTS-outdoor as our testing datasets for synthetic image dehazing. We also use two real-world datasets: Dense-Haze~\cite{ancuti2019dense} and NH-Haze2~\cite{ancuti2021ntire} for real image dehazing.

\textbf{Competitors and Evaluation Metrics.} We compare our method with the prior-based method (\eg, DCP~\cite{he2010single}), physical model based methods(\eg, DehazeNet~\cite{cai2016dehazenet}, AOD-Net~\cite{li2017aod}, and DM2F-Net~\cite{Deng2019}), and hazy-to-clear image translation based methods (\eg, GDN~\cite{liu2019griddehazenet}, GCANet~\cite{chen2019gated}, FFA-Net~\cite{qin2020ffa}, MSBDN~\cite{dong2020multi}, AECR-Net~\cite{wu2021contrastive}, MAXIM-2S~\cite{tu2022maxim}, DeHamer~\cite{guo2022image}, and UDN~\cite{hong2022uncertainty}). We utilize the peak signal-to-noise ratio (PSNR) and structural similarity (SSIM) to evaluate the performance.
\begin{figure*}[t]
	\centering
	\setlength{\abovecaptionskip}{0cm}
	\setlength{\fboxrule}{1pt}
	\setlength{\fboxsep}{0pt}
	\setlength{\tabcolsep}{0.05em}	
	\begin{tabular}{cccccccc}
		PSNR / SSIM& 11.56 / 0.4480  &17.81 / 0.5828 & 19.35 / 0.6666 & 19.83 / 0.6114 & 22.82 / 0.6312 & 22.94 / 0.6776 & $\infty$ / 1 \\	   		
		\includegraphics[width=.12\linewidth]{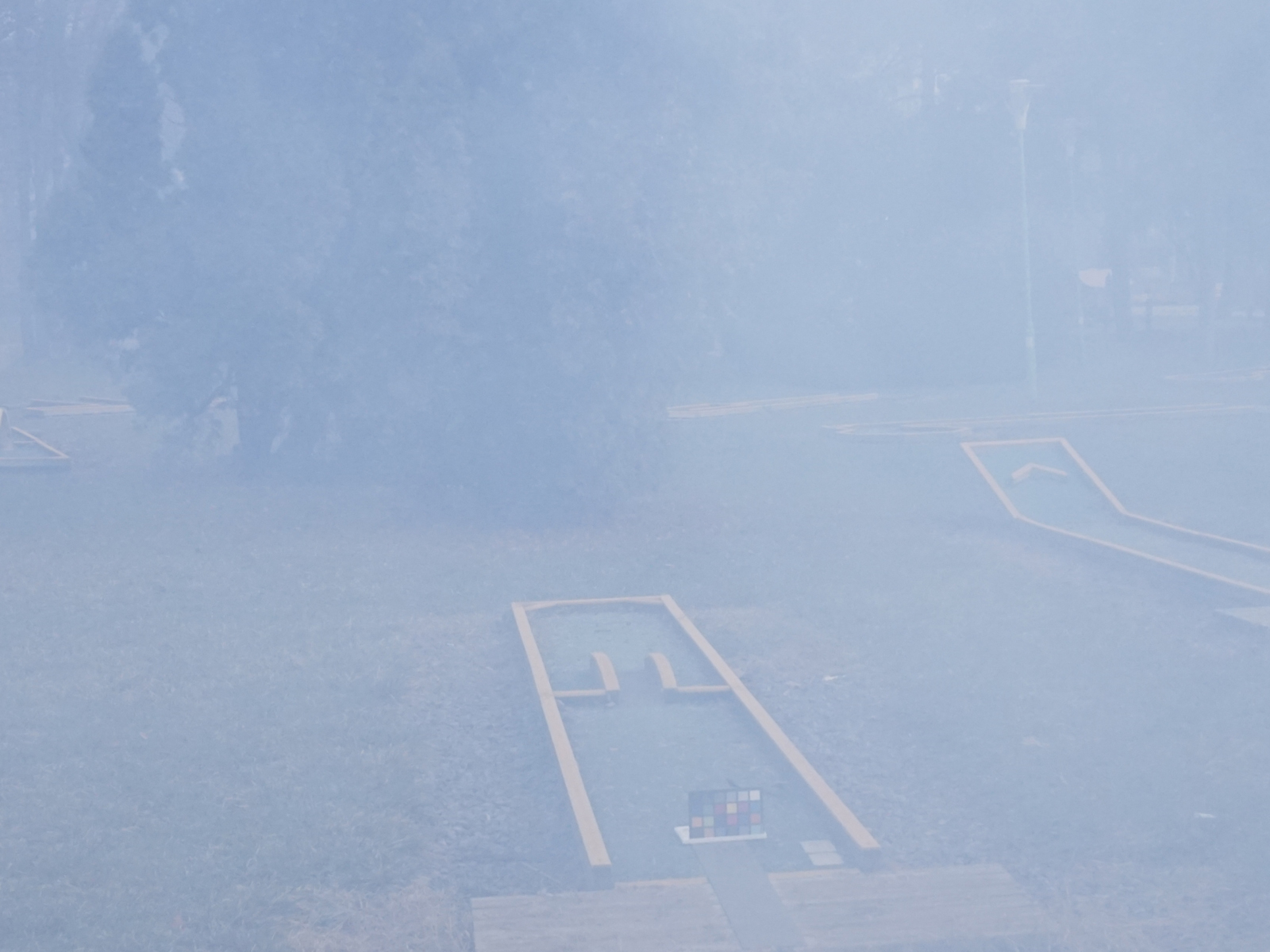} &
		\includegraphics[width=.12\linewidth]{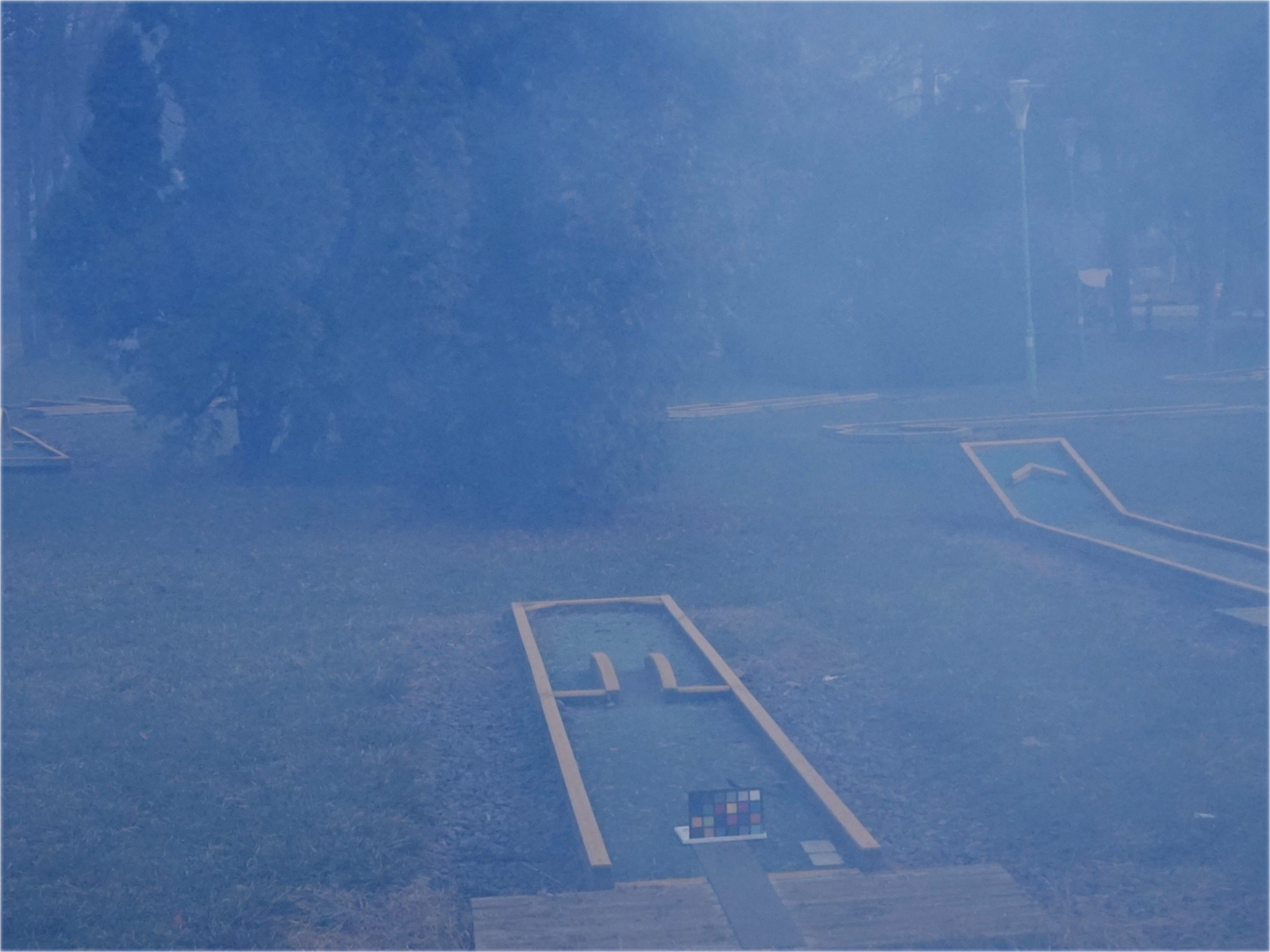} &
		\includegraphics[width=.12\linewidth]{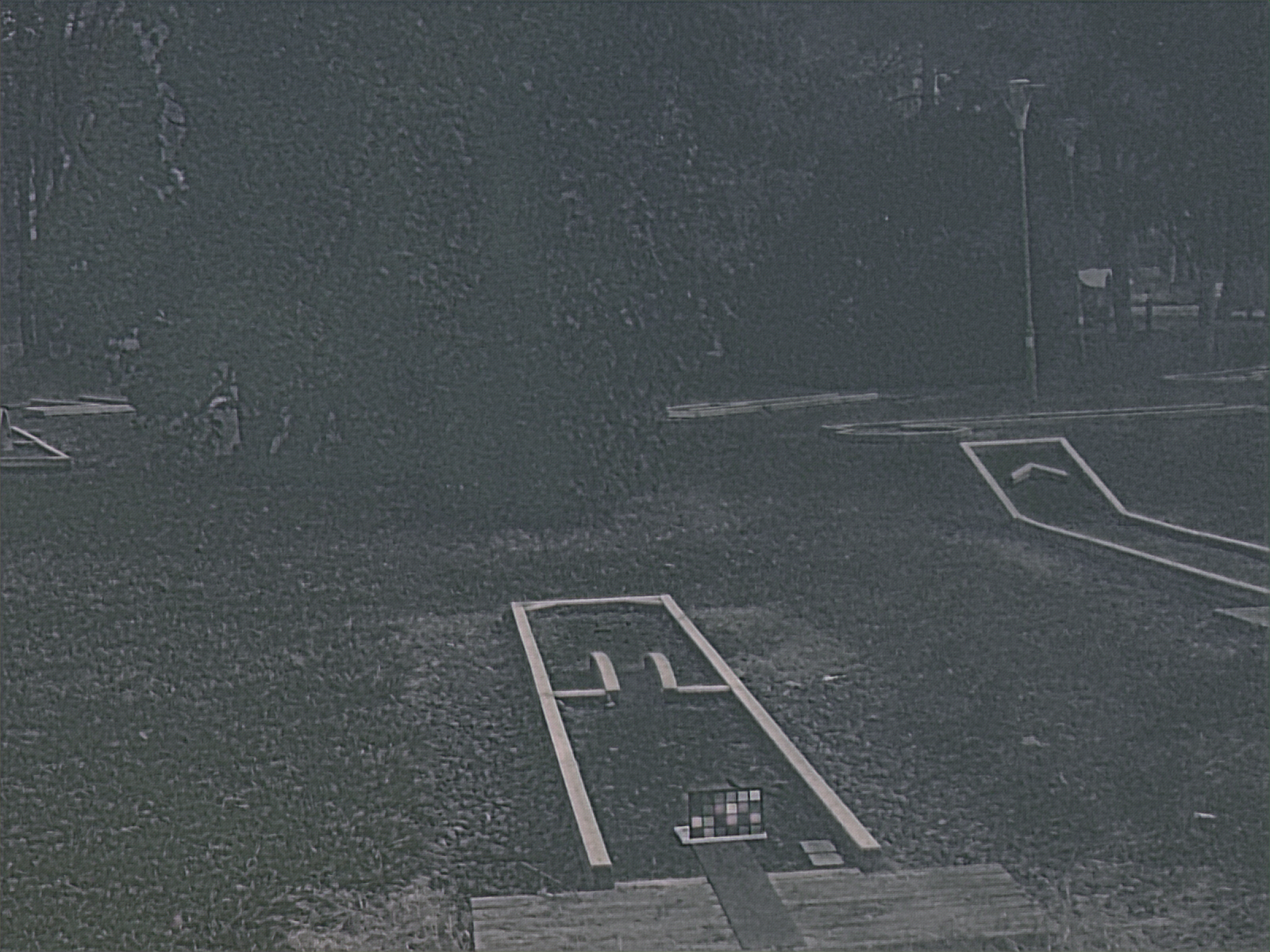} &
		\includegraphics[width=.12\linewidth]{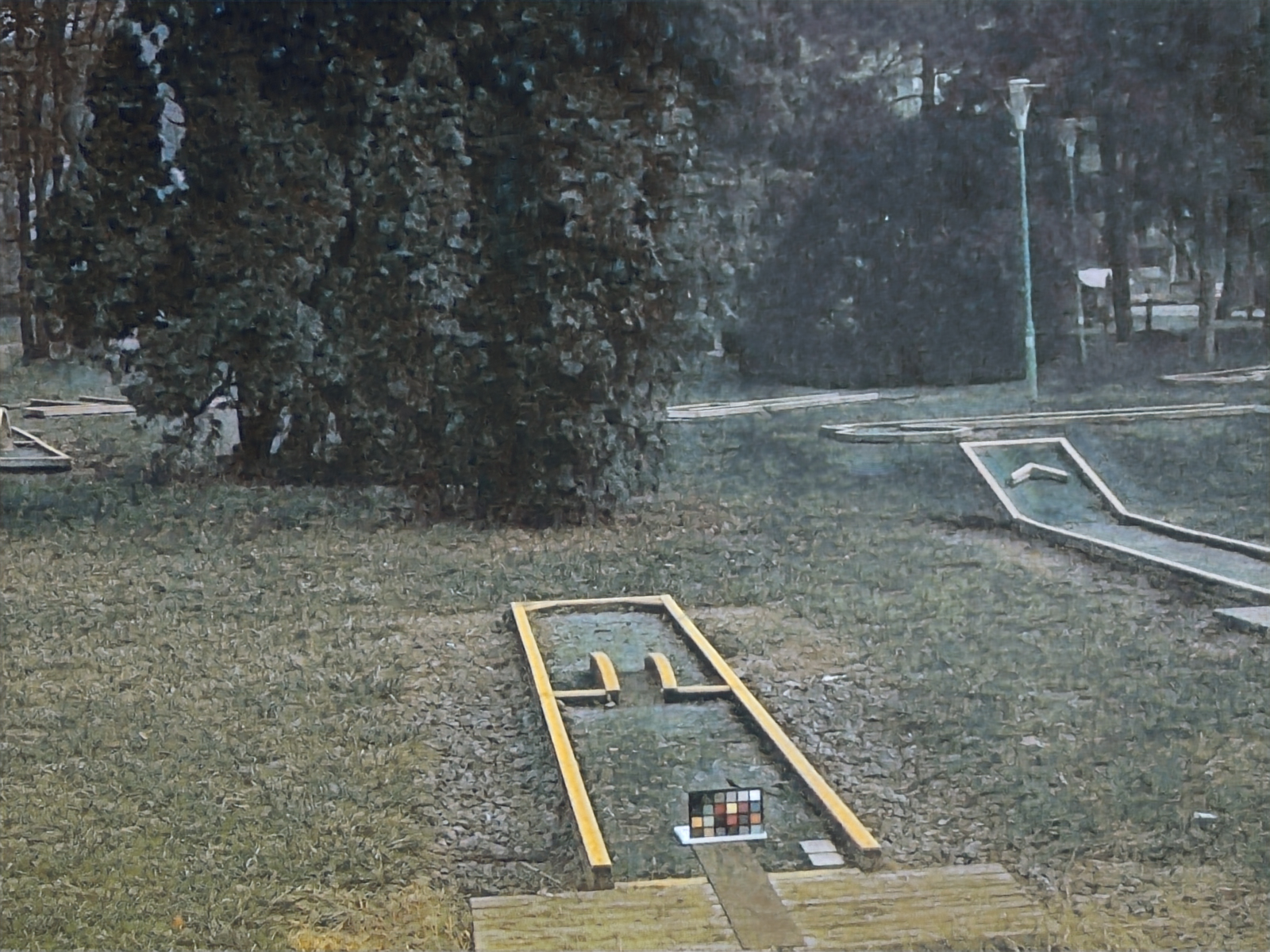} &
		\includegraphics[width=.12\linewidth]{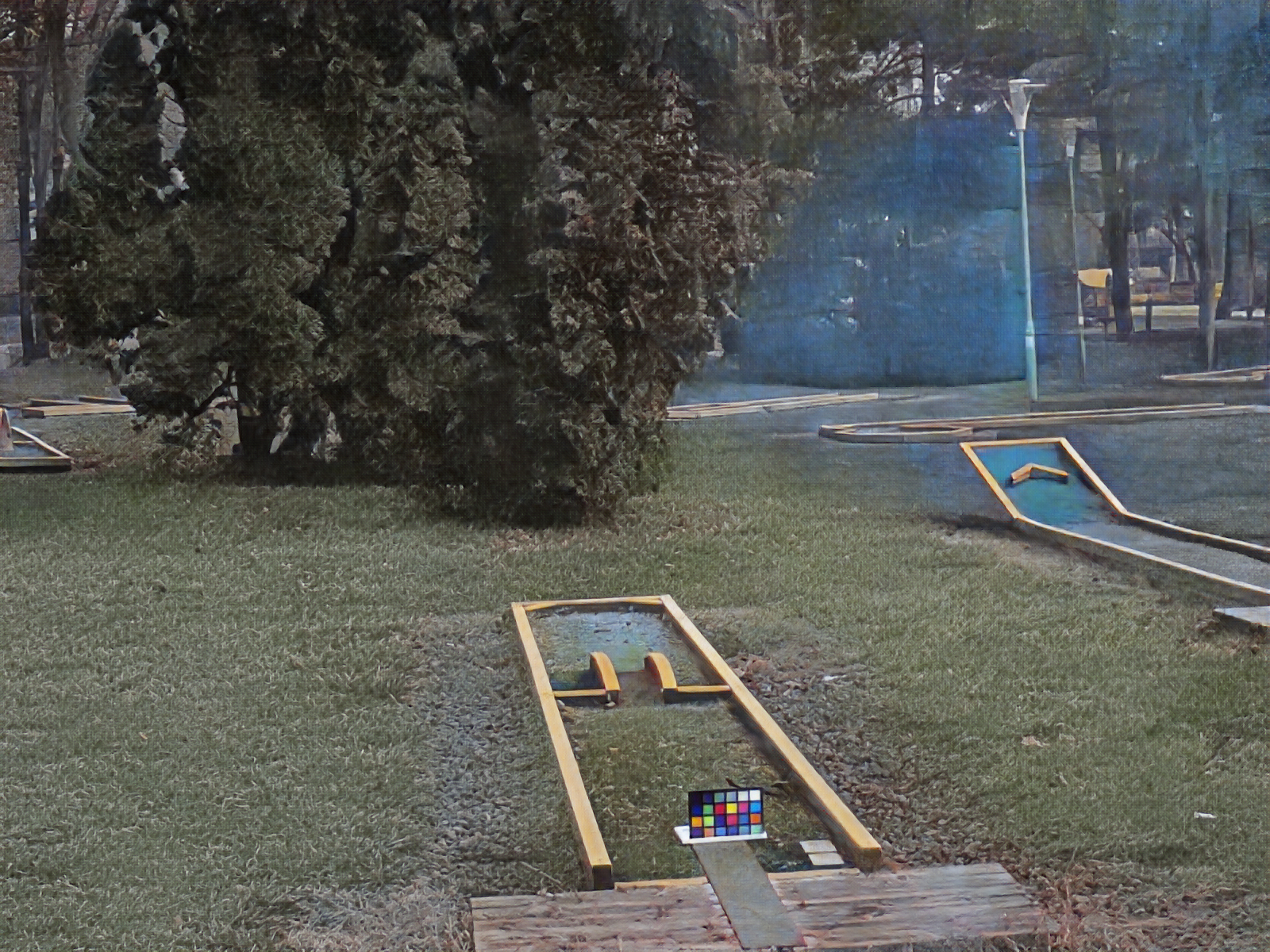} &
		\includegraphics[width=.12\linewidth]{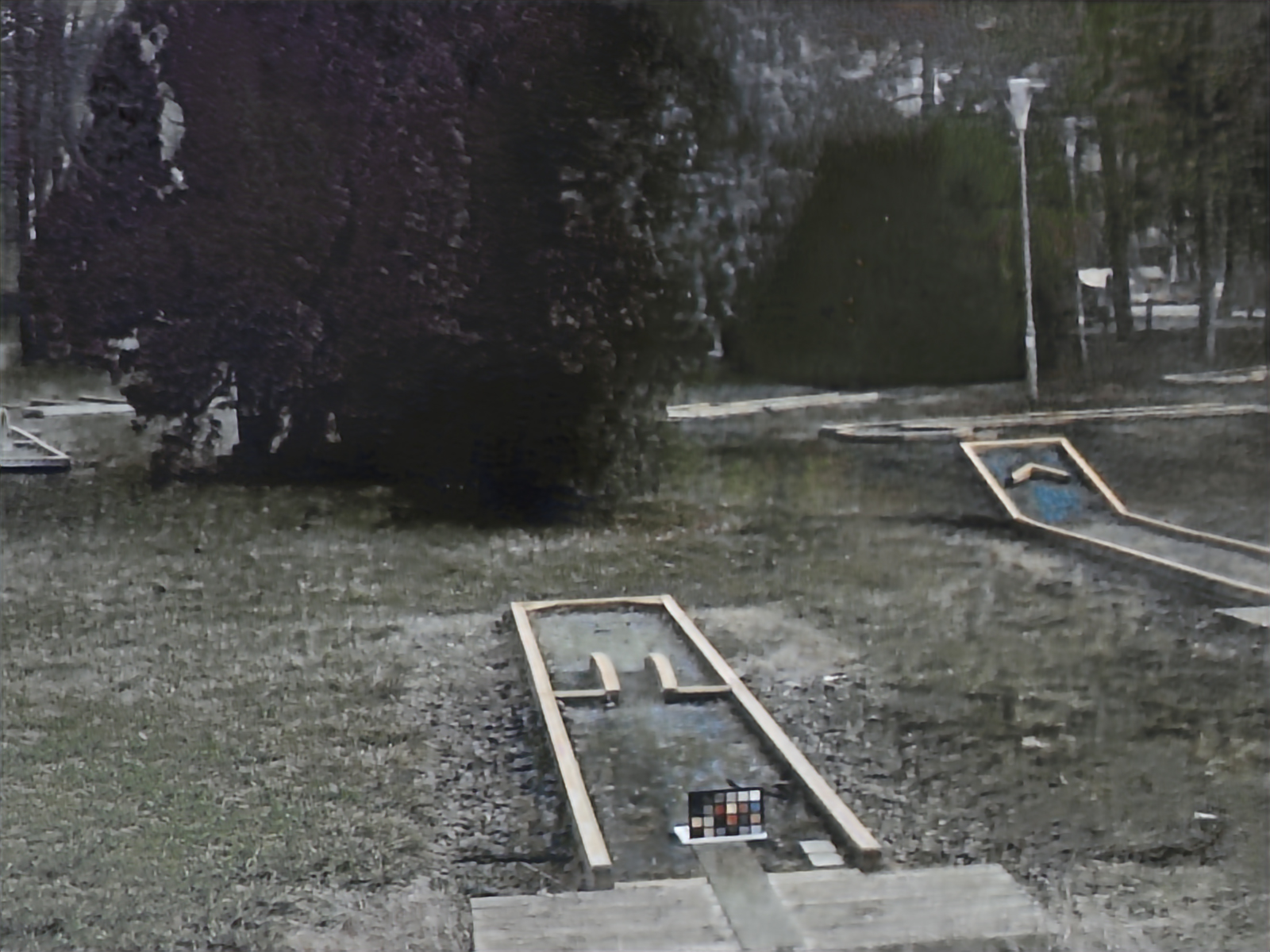} &
		\includegraphics[width=.12\linewidth]{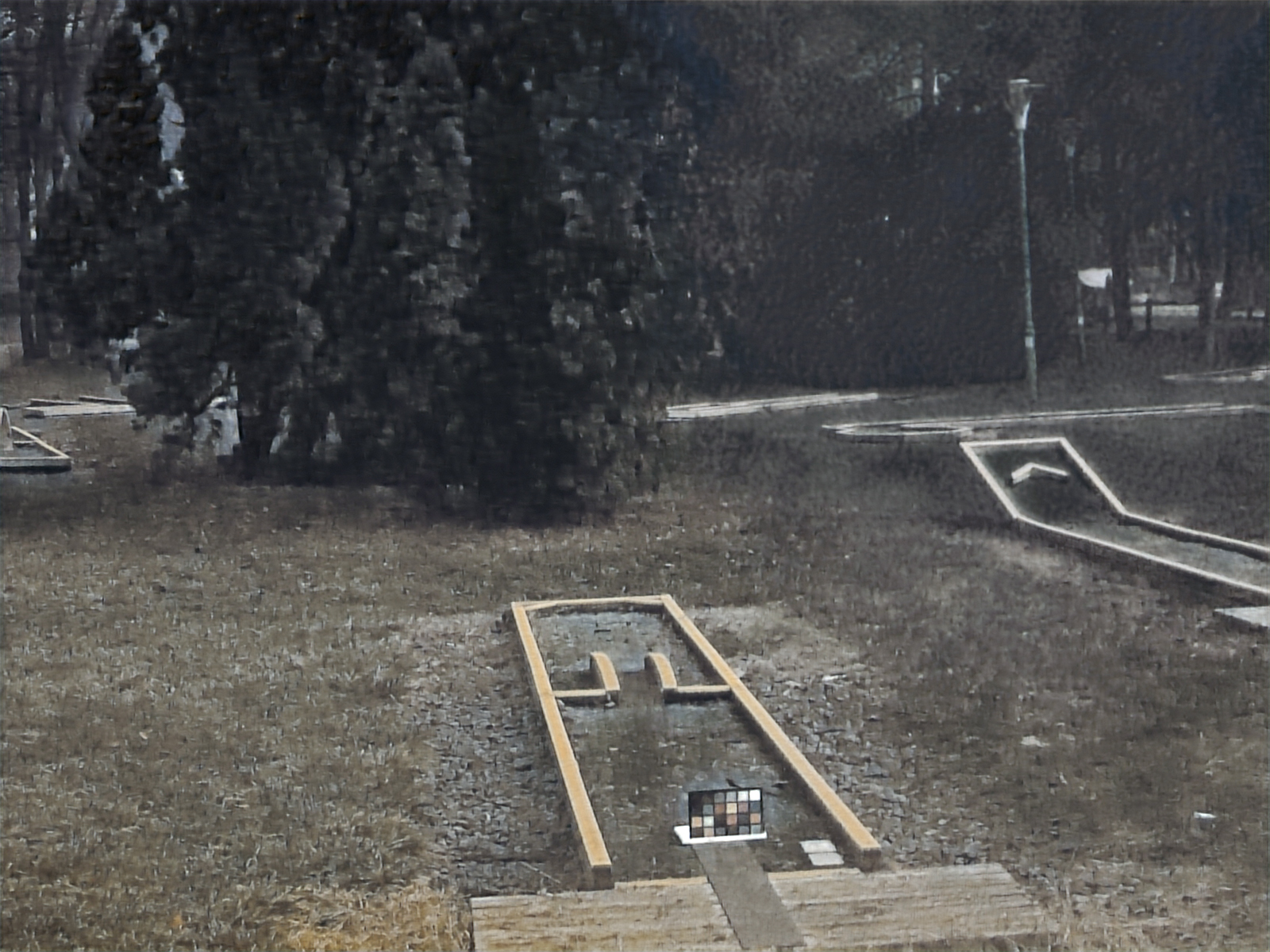}&
		\includegraphics[width=.12\linewidth]{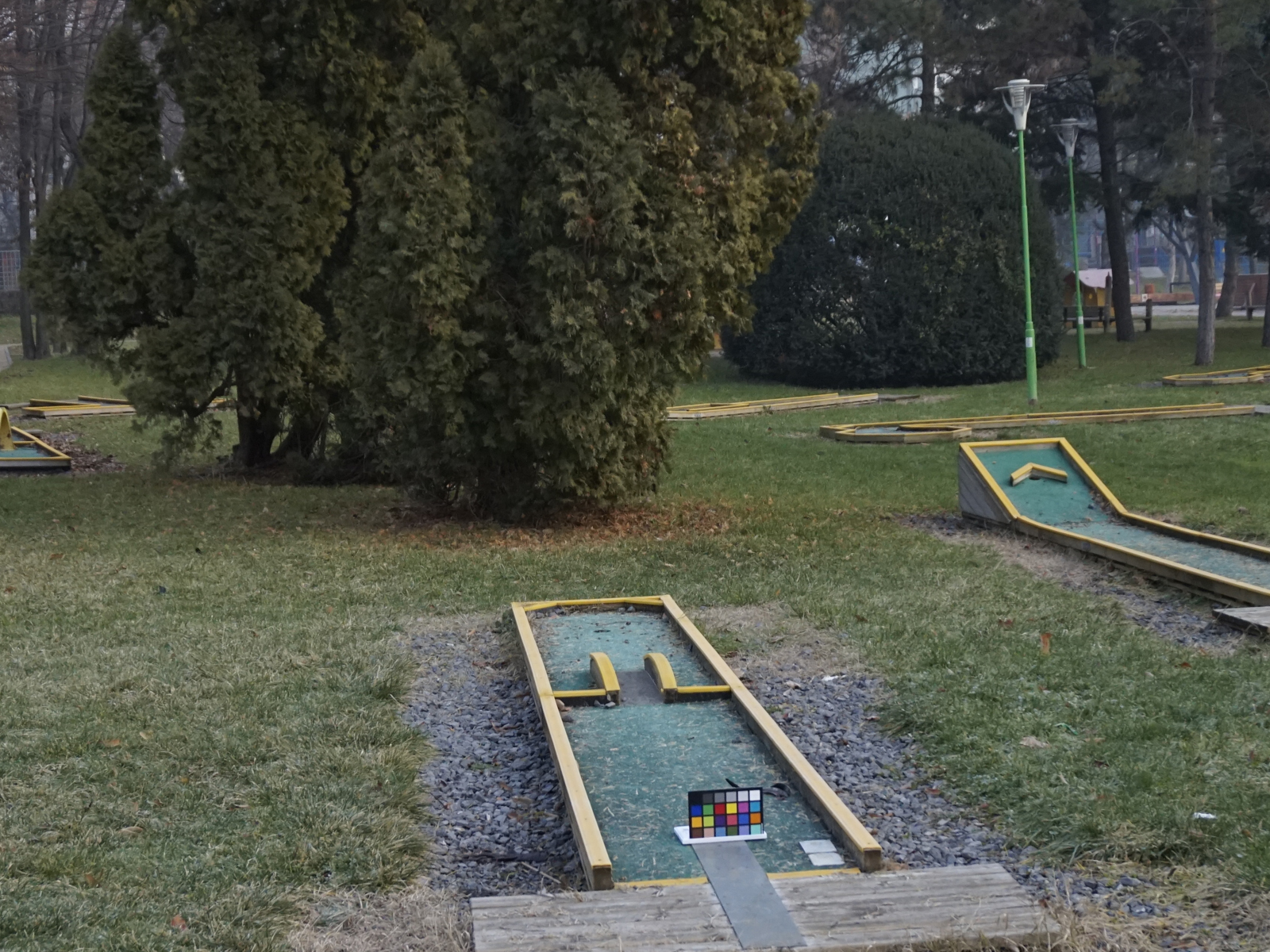}\\			
	\end{tabular}	
	\label{fig:NH19}
	
	\centering
	\setlength{\tabcolsep}{0.05em}	
	\begin{tabular}{cccccccc}
		PSNR / SSIM& 12.22 / 0.5895  &19.30 / 0.7741 & 18.09 / 0.8145 & 19.74 / 0.8312 & 17.21 / 0.7673 & 20.09 / 0.8281 & $\infty$ / 1 \\	   		
		\includegraphics[width=.12\linewidth]{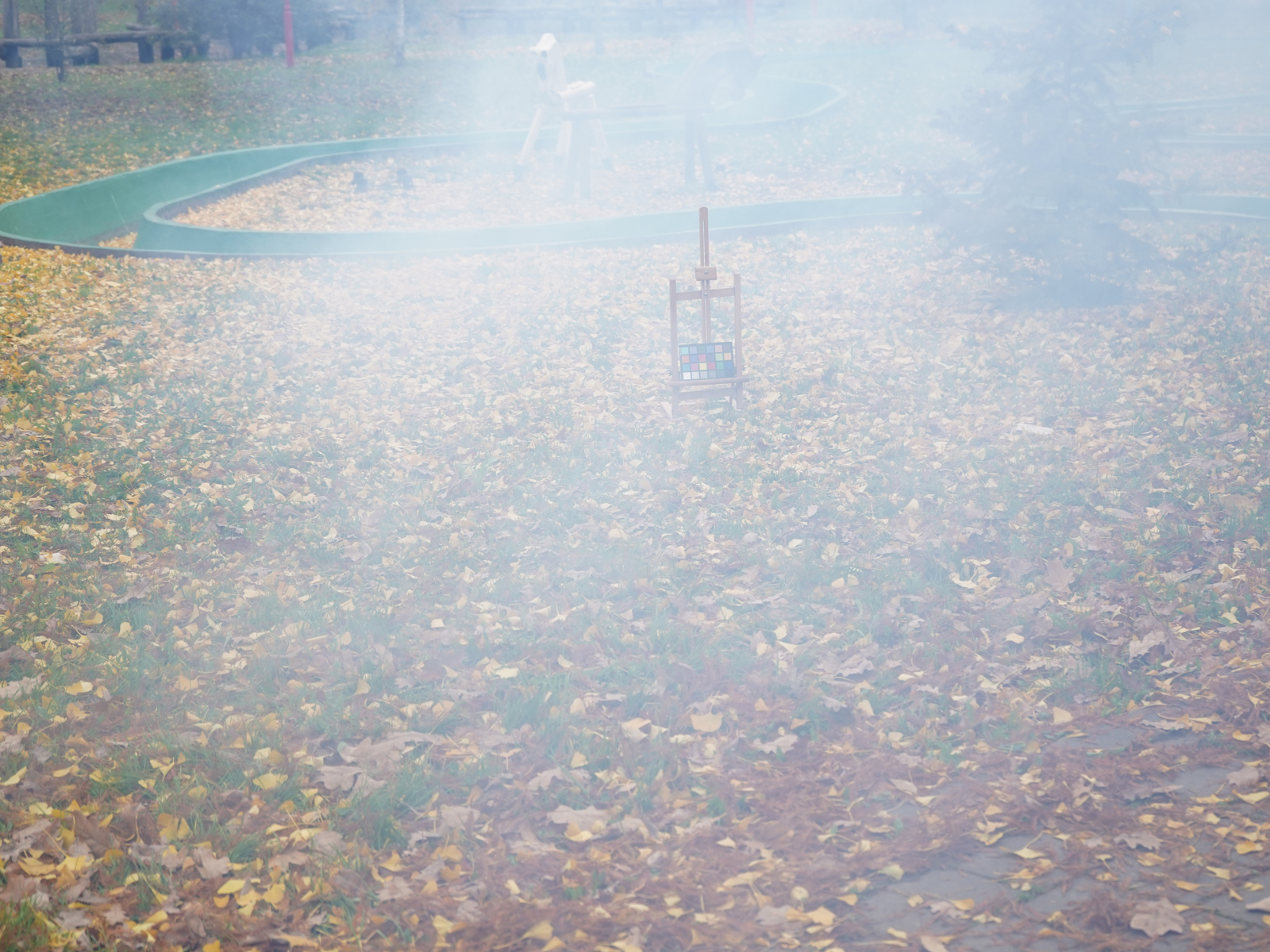} &
		\includegraphics[width=.12\linewidth]{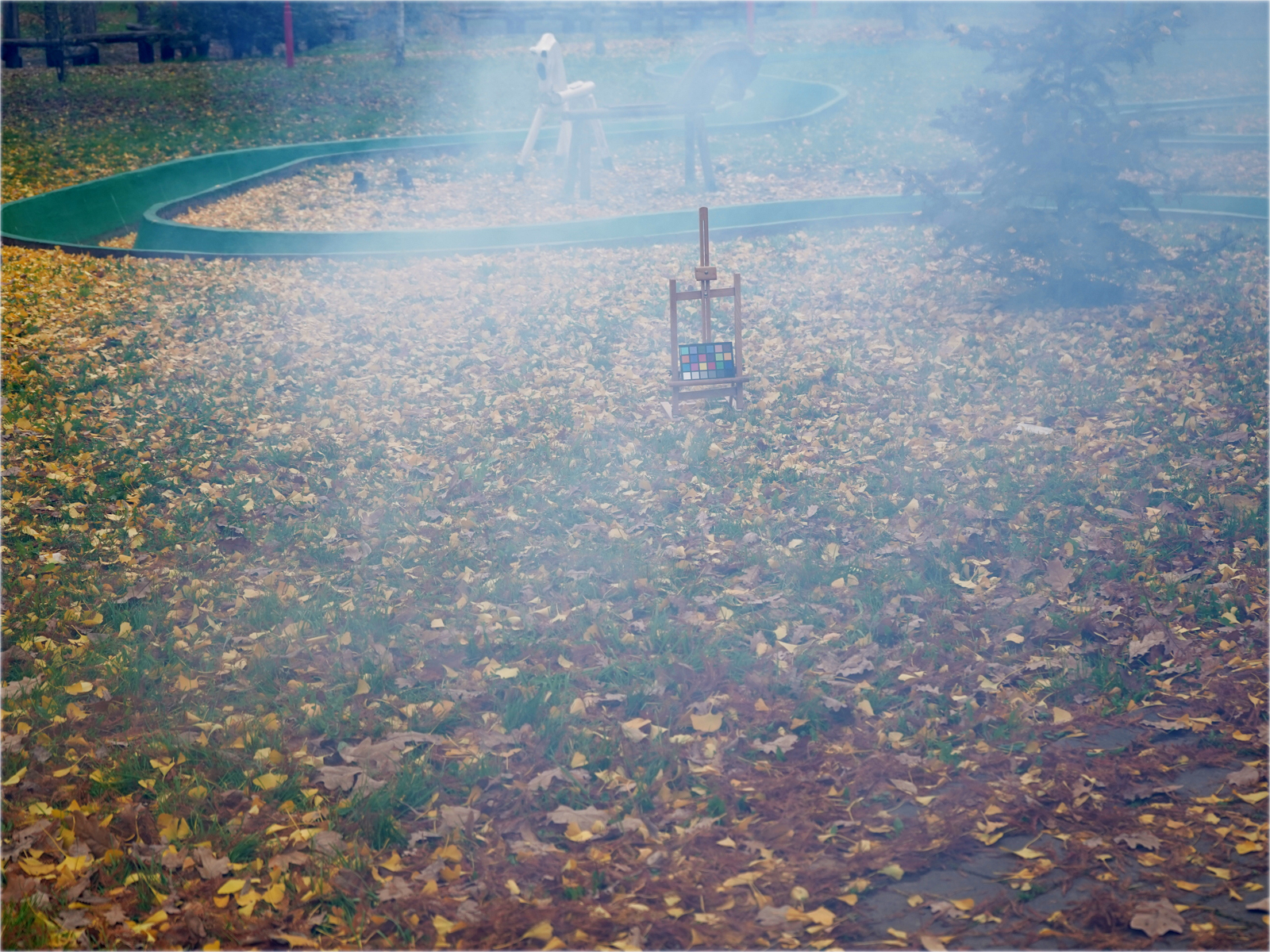} &
		\includegraphics[width=.12\linewidth]{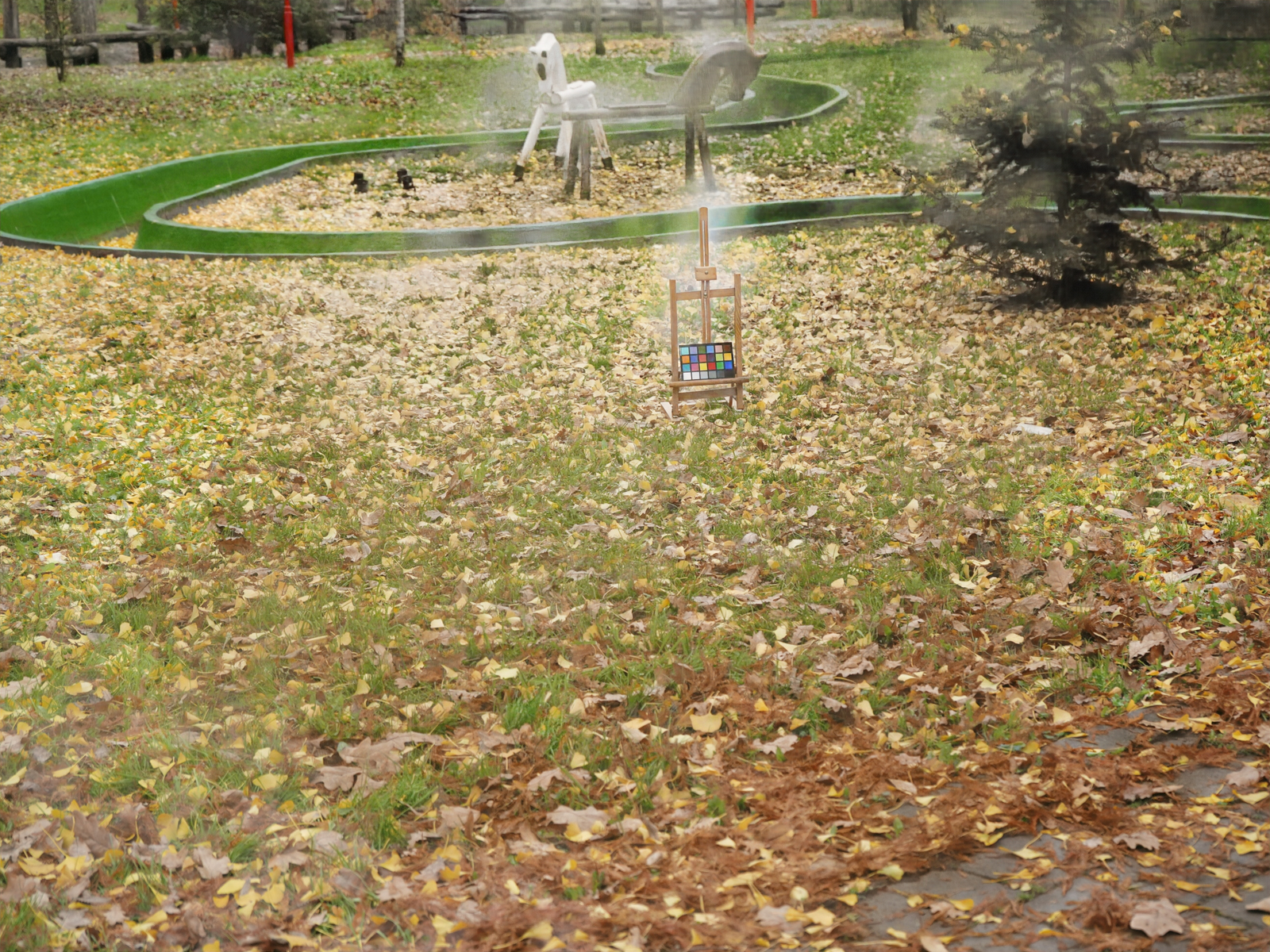} &
		\includegraphics[width=.12\linewidth]{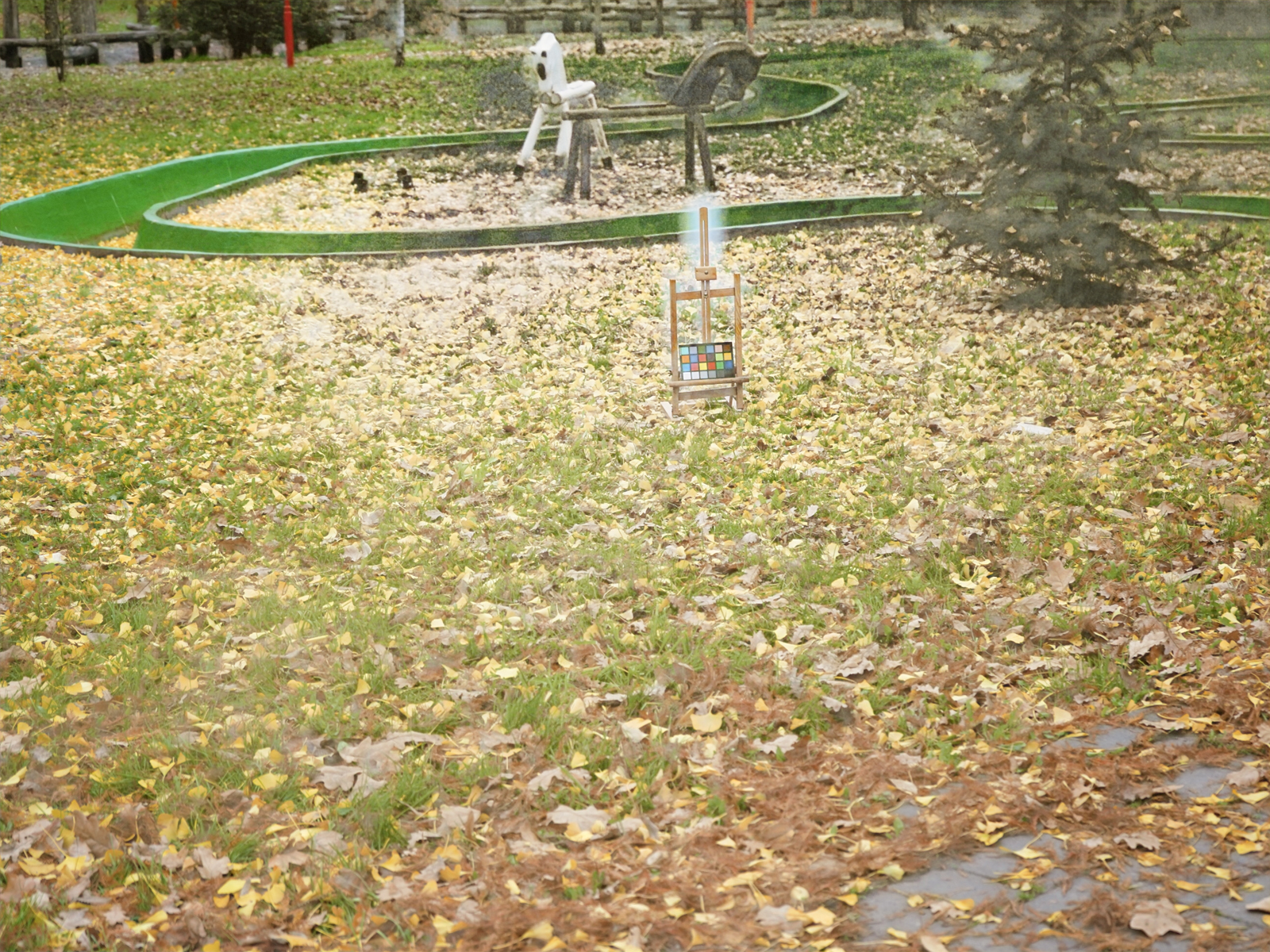} &
		\includegraphics[width=.12\linewidth]{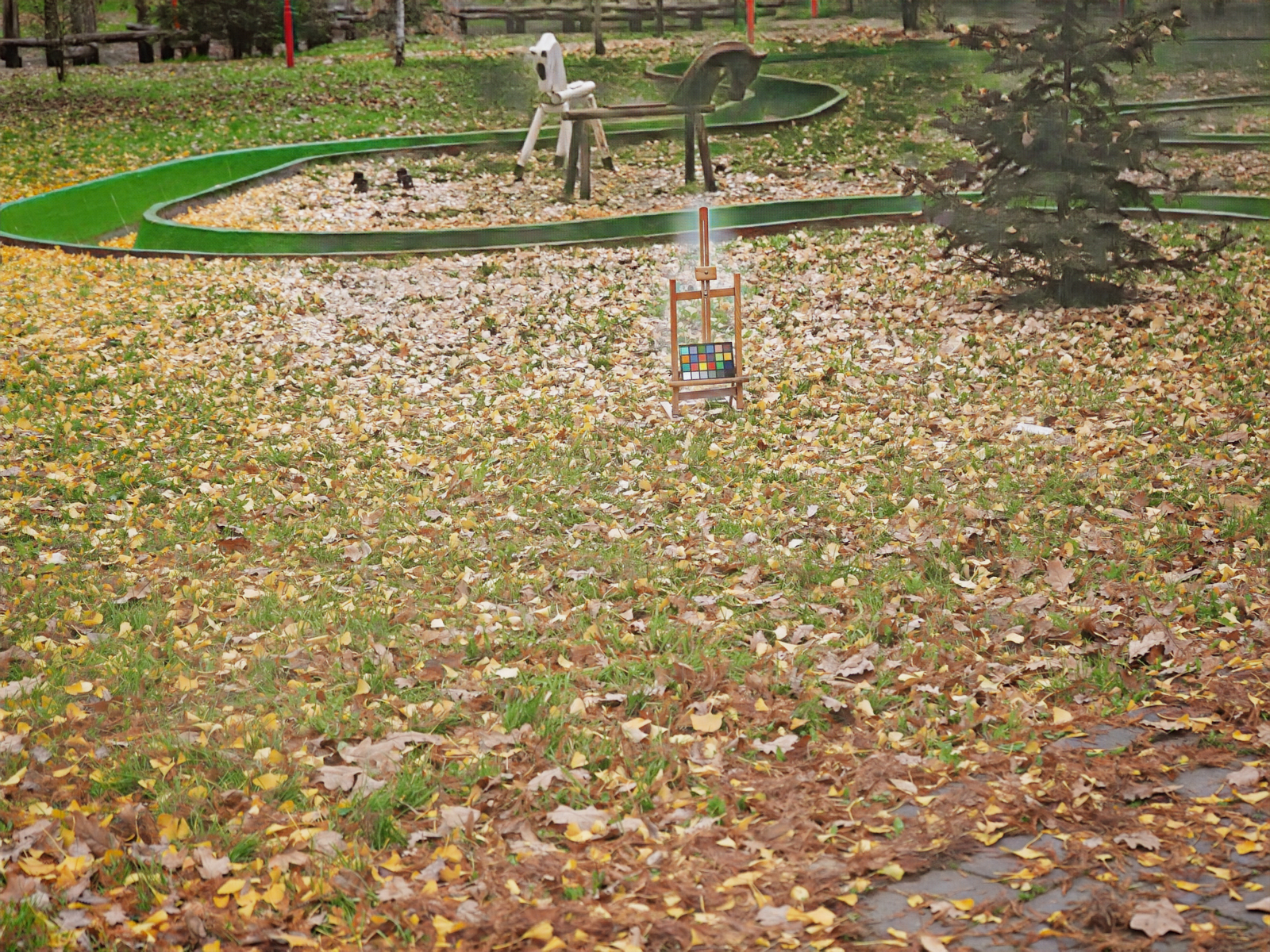} &
		\includegraphics[width=.12\linewidth]{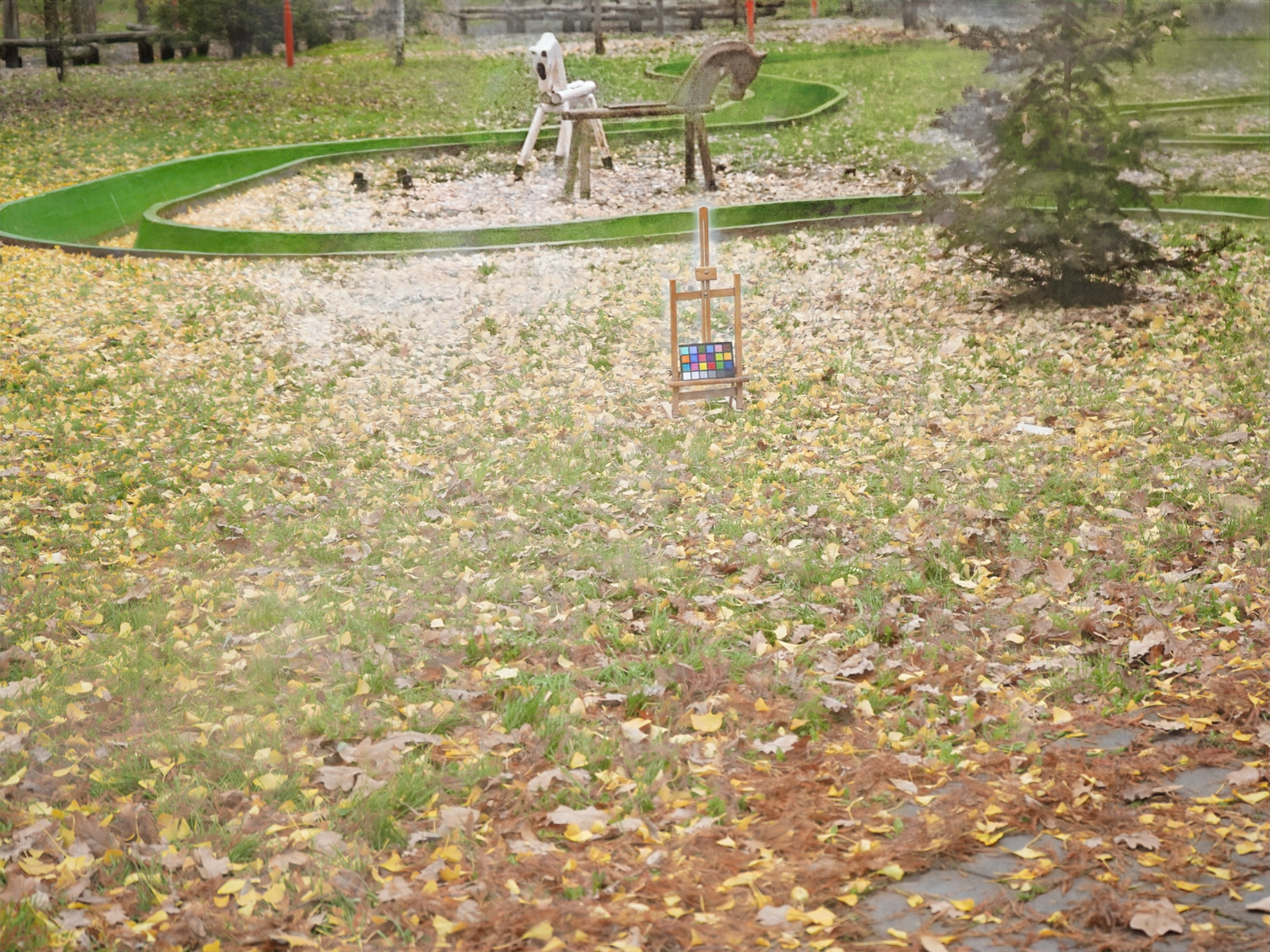} &
		\includegraphics[width=.12\linewidth]{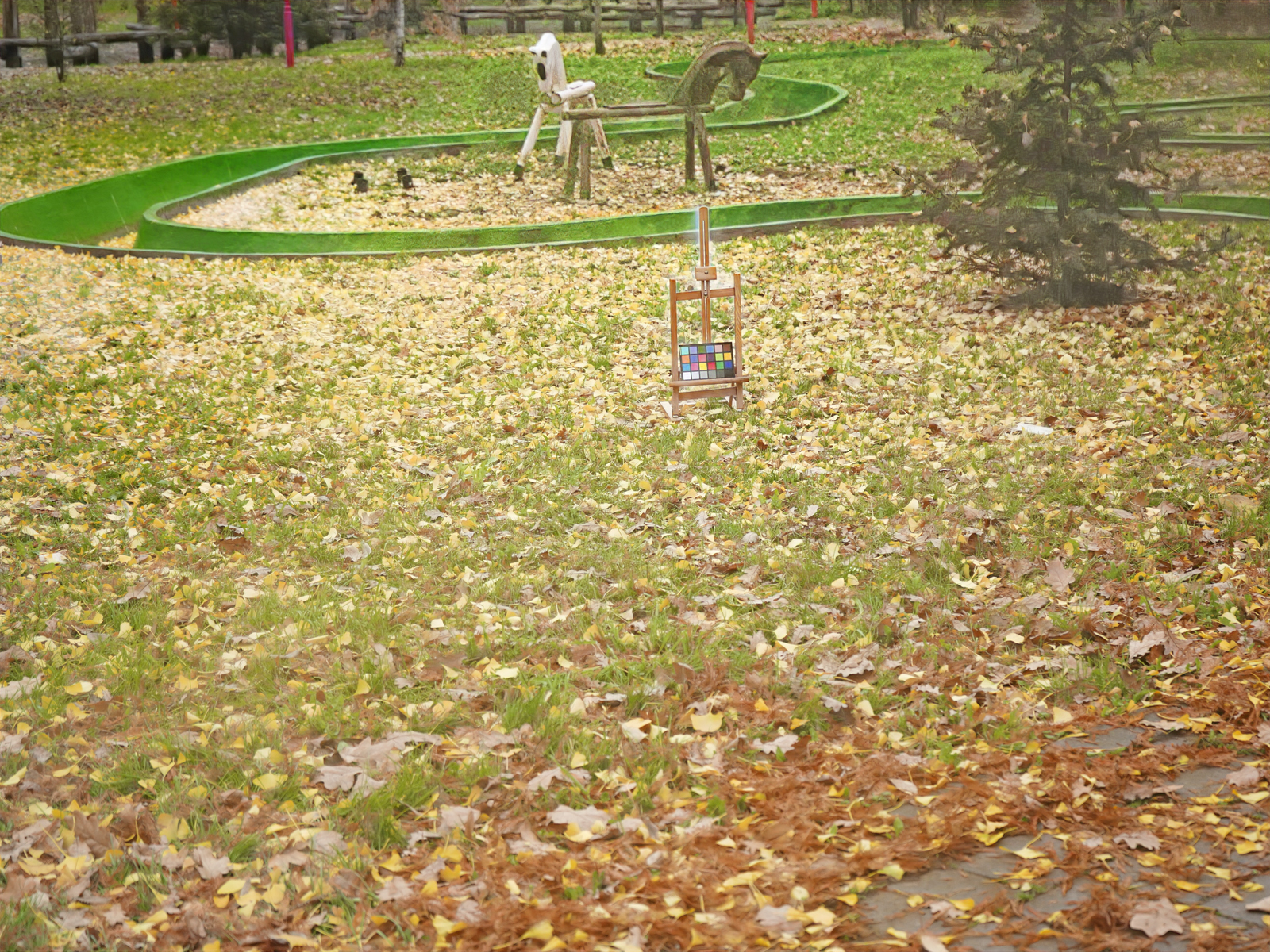}&
		\includegraphics[width=.12\linewidth]{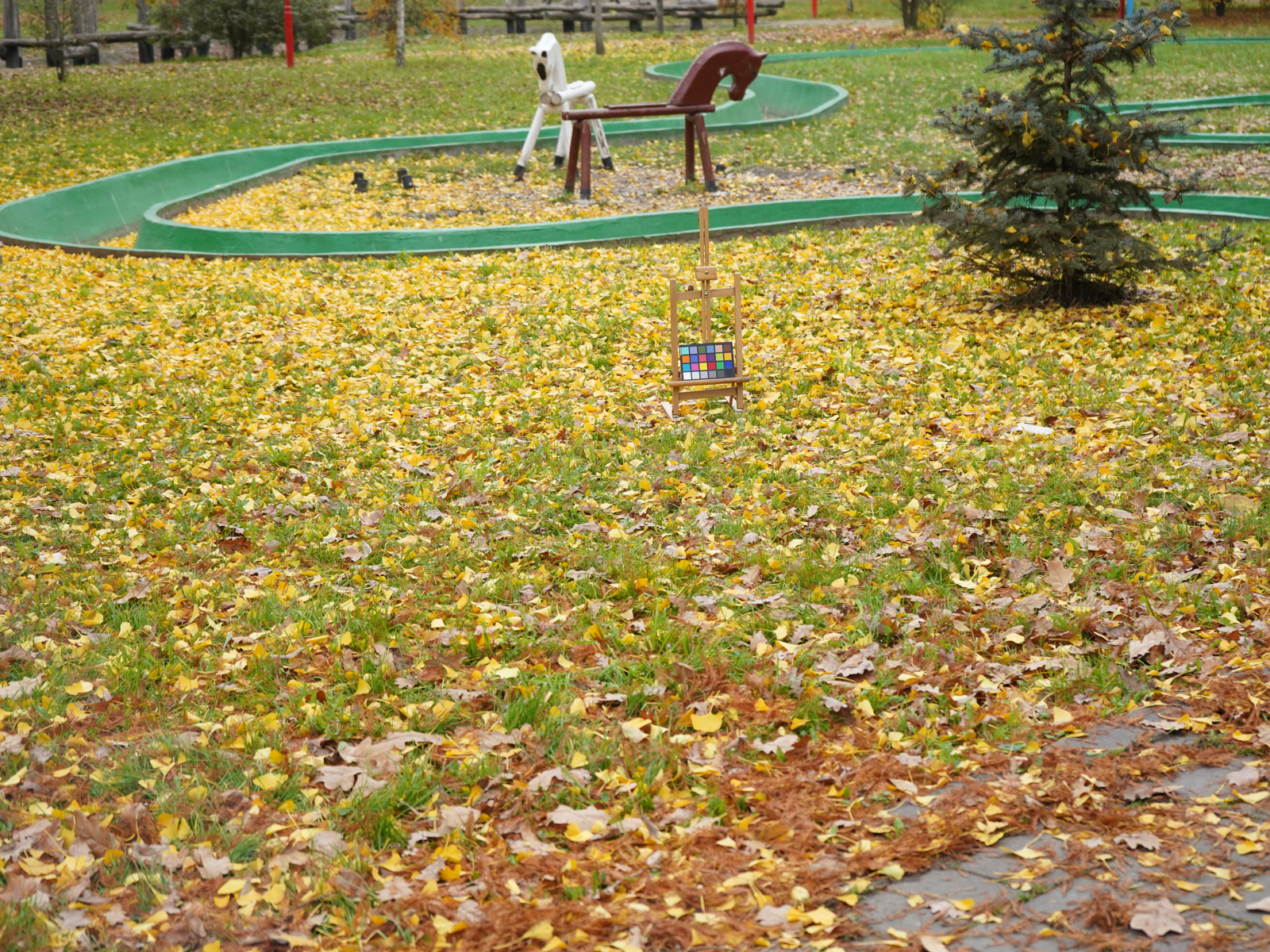}\\			
		Hazy Image &AODNet~\cite{li2017aod}&GDN~\cite{liu2019griddehazenet}&FFA-Net~\cite{qin2020ffa}&AECRNet~\cite{wu2021contrastive}&DeHamer~\cite{guo2022image}&C$^2$PNet (Ours)&GT
	\end{tabular}	
	\caption{Visual results of Dense-Haze (top) and NH-Haze2 (bottom) datasets by different methods. (Zoom in for better view.)}\vspace{-2mm}
	\label{fig:NH21}
\end{figure*}

\subsection{Comparison with SOTAs}
\textbf{Results on Synthetic Datasets.}
Regarding the evaluation of synthetic datasets, Table.~\ref{tab:quantitative} reports the average PSNR and SSIM values of different competitors for SOTS-indoor and SOTS-outdoor datasets. Our C$^2$PNet achieves the best performance on both datasets compared to other SOTAs, with 42.56dB PSNR and 0.9954 SSIM in SOTS-indoor, and 36.68dB PSNR and 0.9900 SSIM in SOTS-outdoor. Specifically, our method outperforms the second-best method UDN by a significant margin on SOTS-indoor, \ie, 3.94dB PSNR and 0.0045 SSIM. Moreover, our method achieves at least 1.50dB PSNR and 0.0029 SSIM performance gains on SOTS-outdoor. In addition, we respectively visualize the recovered images from the SOTS-indoor and the SOTS-outdoor datasets by different methods in Fig.~\ref{fig:indoor} and Fig.~\ref{fig:outdoor}. It can be observed that AODNet and GDN fail to remove most of the haze, while FFA-Net, MAXIM-2S, and DeHamer suffer from severe color distortion, and their results still contain some artifacts. Instead, our method generates the most natural restoration that preserves more details and involves fewer color distortions. Note that we can adjust the number of blocks in our network to balance the performance and the number of parameters. More details are included in the supplementary.

\textbf{Results on Real-world Datasets.}
We also evaluate the proposed C$^2$PNet on real-world datasets including Dense-Haze and NH-Haze2 datasets, summarizing the quantitative results in Table~\ref{tab:quantitative}. It is worth noting that removing haze from real-world images is much more challenging than from synthetic images. Nevertheless, our method outperforms all the other competitors on both datasets in terms of PSNR and SSIM. We also visualize the results in Fig.~\ref{fig:NH21}. Despite the reconstructions of all the comparisons generally being far from good, our method produces the most desired image that succeeded in removing most of the haze. 

\vspace{-2mm}\subsection{Ablation Study}\vspace{-2mm}
\begin{table}[t]
	\caption{Ablation study on C$^2$PNet with different modules and regularizations on SOTS-indoor dataset.}
	\centering
	\small
	\begin{tabular}{c||c|c}
		\toprule
		Model&PSNR&SSIM\\	
		\midrule
		base (FFA-Net) &36.39&0.9886\\
		
		base+FDU&36.59&0.9894\\
		
		base+PDU  &38.30&0.9914\\		
		
		base+PDU+CR(non-consensual,1:10)  &41.32&0.9947\\	
		
		base+PDU+CR(consensual,1:7)+w/o CL  &42.09&0.9951\\		
		\midrule
		\textbf{Ours (1:7)} &\textbf{42.56}&\textbf{0.9954}\\
		\bottomrule
	\end{tabular}\vspace{-5mm}
	\label{tab:ablation}
\end{table}

\begin{table*}[t]
	\caption{Evaluation of applying curricular contrastive regularization into SOTAs.}\vspace{-2mm}
	\centering	
	\resizebox{0.85\textwidth}{!}{
		\begin{tabular}{c|c|c|c||c||c|c|c|c|c}
			\toprule
			\multicolumn{4}{c||}{Regularization}&\multirow{2}*{Metric}&\multicolumn{5}{c}{Method}\\
			\cmidrule(lr){1-4}		
			\cmidrule(lr){6-10}
			CR&Space&CL&Rate&&GCANet~\cite{chen2019gated}&GDN~\cite{liu2019griddehazenet}&MSBDN~\cite{dong2020multi}&FFANet~\cite{qin2020ffa}&DMTNet~\cite{liu2021synthetic}\\
			\midrule
			\multirow{2}*{\XSolidBrush}&\multirow{2}*{N/A}&\multirow{2}*{N/A}&\multirow{2}*{N/A}&PSNR&30.06&32.16&33.79&36.39&28.53\\
			
			&&&&SSIM&0.9596&0.9836&0.9835&0.9886&0.96\\
			\midrule
			\multirow{2}*{\Checkmark}&\multirow{2}*{\footnotesize{non-consensual}}&\multirow{2}*{\XSolidBrush}&\multirow{2}*{1:10}&PSNR&29.83&33.36&34.74&37.21&30.88\\
			&&&&SSIM&0.9611&0.9867&0.9859&0.9920&0.9785\\
			\midrule
			\multirow{2}*{\Checkmark}&\multirow{2}*{\footnotesize{consensual}}&\multirow{2}*{\XSolidBrush}&\multirow{2}*{1:7}&PSNR&29.91&34.91& 34.95&38.93&31.16\\
			&&&&SSIM&0.9612&\textbf{0.9892}&0.9865&0.9936&0.9772\\
			\midrule
			\multirow{2}*{\Checkmark}&\multirow{2}*{\footnotesize{consensual}}&\multirow{2}*{\footnotesize{self-paced}}&\multirow{2}*{1:7}&PSNR&30.05&35.20&35.17&38.98&31.56\\
			&&&&SSIM&0.9596&0.9889&0.9861&0.9936&0.9776\\
			\midrule
			\multirow{2}*{\Checkmark}&\multirow{2}*{\footnotesize{\textbf{consensual}}}&\multirow{2}*{\footnotesize{\textbf{ours}}}&\multirow{2}*{\textbf{1:7}}&PSNR&\textbf{30.76}&\textbf{35.46}&\textbf{35.31}&\textbf{39.24}&\textbf{31.63}\\
			&&&&SSIM&\textbf{0.9668}&0.9880&\textbf{0.9875}&\textbf{0.9937}&\textbf{0.9791}\\ 		
			\bottomrule
	\end{tabular}}\vspace{-2mm}
	\label{tab:generality}
\end{table*}
In this section, we analyze the effectiveness of the different components of the proposed C$^2$PNet, including PDU, consensual negatives-based contrastive regularization (consensual CR), and curricular contrastive regularization (C$^2$R). Our \textit{base} network is FFA-Net, and subsequently, we establish five variants including 1) \textbf{base+FDU}: Replacing the PA module with FDU in the FA block. 2) \textbf{base+PDU}: Replacing the PA module with PDU in the FA block. 3) \textbf{base+PDU+CR(non-consensual, 1:10)}: Adding canonical contrastive regularization to base+PDU, with the rate between positive and negative samples being 1:10. 4) \textbf{base+PDU+CR(consensual, 1:7)+w/o CL}: Adding consensual CR without our curriculum strategy (CL) to base+PDU, with the rate between positive and negative samples being 1:7. 5) \textbf{Ours}: The full model of our C$^2$PNet. We list the results in Table~\ref{tab:ablation}, using the ITS dataset for training and SOTS-indoor for testing.

\textbf{Effectiveness of PDU.} The architecture of PDU is derived from Eq.~\eqref{equ:final} with a consideration of the physical characteristics of $A$ and $T$, which introduces a dual-branch interaction for the prediction of both factors. Since the features corresponding to $A$ and $T$ are disentangled by our PDU, the latent structural feature-level information is excavated more accurately. As a result, in Table~\ref{tab:ablation} we can see that the PDU achieves 1.71dB and 1.91dB gains over base+FDU and the base network, respectively. 

\textbf{Effectiveness of consensual CR.} 
We follow the same setting as non-consensual CR that considers at most 10 negatives due to the practicability towards training time and GPU memory limitations, and we use the optimal numbers of negatives for a fair comparison, \ie, 7 (consensual CR) vs. 10 (non-consensual CR). It can be observed that consensual CR remarkably boosts the performance against base+PDU and base+PDU+CR (non-consensual, 1:10) with PSNR improvements of 3.79dB and 0.77dB, respectively. Note that our training time is accelerated to 137 hours in contrast to 200 hours for non-consensual CR (1:10). These facts reinforce the superiority of consensual CR. More analysis can be found in the supplementary. 


\textbf{Effectiveness of C$^2$R.} Our full network employs the proposed CL strategy into consensual CR during training and performs the best in comparison with all the variants. Compared to base+PDU+CR(consensual, 1:7)+w/o CL, C$^2$PNet achieves an increase of 0.57dB in PSNR, revealing the effectiveness of the proposed C$^2$R.

\subsection{Generality Analysis for C$^2$R}
\vspace{-2mm}To further verify the generality of our C$^2$R, we apply it to different SOTA methods and compare it with several other universal regularizations. The results are summarized in Table~\ref{tab:generality}. Our method achieves significant improvements in PSNR and SSIM on all five SOTAs compared to other regularizations, except for a slight decrease of 0.0012 in SSIM compared to consensual CR on GDN. Specifically, our C$^2$R enhances the performances of the five baseline models with average PSNR improvements of 0.70-3.30dB, and is superior to CR (non-consensual,1:10) as a regularization term by average PSNR improvements of 0.93-2.10dB. In particular, compared to the popular self-paced CL strategy~\cite{Kumar2010}, our CL method yields a maximum increase of 0.71dB in PSNR. The possible reason is that using the self-paced strategy will feed the negatives into the regularization stage by stage, leading to 1) a two-level split of difficulty without considering the ultra-hard negatives and 2) all the introduced negatives share the same weight. However, as we analyzed before, both hard and ultra-hard samples can provide useful information for regularization during training, and the corresponding weights need to be delicately assigned separately.

\vspace{-2mm}\section{Discussion and Limitation}\vspace{-2mm}
An important advantage of negatives from existing dehazing models is the post-dehazing priors embedded in the recoveries, such as the distribution of the haze residue, which can indicate a more challenging pattern that is difficult to remove. This can provide valuable information to the model during training. However, as most existing methods perform poorly in real-world scenarios, it is hard to collect high-quality images as the non-easy (especially ultra-hard) negatives. This may limit the capacity of our model, despite achieving promising performance on real-world dehazing.

\vspace{-2mm}\section{Conclusion}\vspace{-2mm}
\label{conclude}
In this paper, we propose a novel C$^2$PNet for single image dehazing. Instead of using non-consensual negatives, we introduce consensual negatives to construct contrastive samples and then apply a curricular contrastive regularization that considers the difficulty of the negatives to constrain a more compact solution space. To enhance the interpretability of the feature space, we further design a physics-aware dual-branch unit based on the physics model. The features produced by the unit are enforced to conform with the hazing process, thus facilitating haze removal. Extensive experiments demonstrate the validity and generality of the proposed method.

\small{\textbf{Acknowledgements:} This project is supported by the National Natural Science Foundation of China (No. 62102381, U1706218, 41927805, 61972162); Shandong Natural Science Foundation (ZR2021QF035); the China Postdoctoral Science Foundation (2021T140631); the National Key R\&D Program of China (2018AAA0100600); Guangdong Natural Science Funds for Distinguished Young Scholar (No. 2023B1515020097); and Guangdong Natural Science Foundation (No. 2021A1515012625).}

{\small
\bibliographystyle{ieee_fullname}
\bibliography{aaai23}
}

\end{document}